\documentclass[conference]{IEEEtran}

\usepackage{CJKutf8}

\usepackage{multirow}
\usepackage{amsmath}
\usepackage{threeparttable}
\usepackage{array}
\usepackage{makecell}
\usepackage{url}
\usepackage{bm}
\usepackage{amsmath}
\usepackage{mathrsfs}
\usepackage{amssymb}
\usepackage{graphicx}
\usepackage{subfigure}
\usepackage{caption}
\usepackage{mathrsfs}
\usepackage{csquotes}
\usepackage{colortbl}

\definecolor{mygray}{gray}{.1}

\newcommand{\PreserveBackslash}[1]{\let\temp=\\#1\let\\=\temp}

\definecolor{mycyan}{cmyk}{.3,0,0,0}
\definecolor{mygray}{gray}{.85}

\newcolumntype{R}[1]{>{\raggedleft\arraybackslash}p{#1}}

\makeatletter
\newcommand{\mypm}{\mathbin{\mathpalette\@mypm\relax}}
\newcommand{\@mypm}[2]{\ooalign{%
  \raisebox{.1\height}{$#1+$}\cr
  \smash{\raisebox{-.6\height}{$#1-$}}\cr}}
\makeatother

\newlength\myequborder
\newenvironment{nospaceflalign*}
 {\setlength{\abovedisplayskip}{\myequborder}\setlength{\belowdisplayskip}{\myequborder}%
  \csname flalign*\endcsname}
 {\csname endflalign*\endcsname\ignorespacesafterend}

\begin{document}
%
\title{A Boundary Regression Model for Nested \\Named Entity Recognition}


\author{\IEEEauthorblockN{Yanping Chen\IEEEauthorrefmark{1},
Lefei Wu\IEEEauthorrefmark{2},
Qinghua Zheng\IEEEauthorrefmark{2},
Ruizhang Huang\IEEEauthorrefmark{1},
Jun Liu\IEEEauthorrefmark{2},\\
Liyuan Deng\IEEEauthorrefmark{1},
Junhui Yu\IEEEauthorrefmark{1},
Yongbin Qing\IEEEauthorrefmark{1},
Bo Dong\IEEEauthorrefmark{2},
Ping Chen\IEEEauthorrefmark{3}}

\IEEEauthorblockA{\IEEEauthorrefmark{1} Guizhou University, Guiyan, China}
\IEEEauthorblockA{\IEEEauthorrefmark{2} Xian'an Jiaotong University, Xi'an, China}
\IEEEauthorblockA{\IEEEauthorrefmark{3} University of Massachusetts Boston, Boston, USA}}


\maketitle

\begin{abstract}
Recognizing named entities (NEs) is commonly conducted as a classification problem that predicts a class tag for a word or a NE candidate in a sentence. In shallow structures, categorized features are weighted to support the prediction. Recent developments in neural networks have adopted deep structures that map categorized features into continuous representations. This approach unfolds a dense space saturated with high-order abstract semantic information, where the prediction is based on distributed feature representations. In this paper, positions of NEs in a sentence are represented as continuous values. Then, a regression operation is introduced to regress boundaries of NEs in a sentence. Based on boundary regression, we design a boundary regression model to support nested NE recognition. It is a multiobjective learning framework, which simultaneously predicts the classification score of a NE candidate and  refine its spatial location in a sentence. It has the advantage to resolve nested NEs and support boundary regression for locating NEs in a sntence. By sharing parameters for predicting and locating, this model enables more potent nonlinear function approximators to enhance model discriminability. Experiments demonstrate state-of-the-art performance for nested NE recognition\footnote{Our codes to implement the BR model are available at: \url{https://github.com/wuyuefei3/BR}.}.
\end{abstract}

\IEEEpeerreviewmaketitle

\section{Introduction}
\label{sec:introduction}

Named entity (NE) recognition is often modeled as a sequence labelling task, where a sequence model (e.g., Conditional Random Field (CRF) \cite{mccallum2003early} or Long Short-Term Memory (LSTM) \cite{hochreiter1997long}) is adopted to output a maximized labelling sequence. Because sequence models are effective to encode semantic dependencies of a sentence and constraint the structure of a labelling sequence, they have achieved a great success for NE recognition. However, sequence models assume a flattened structure for each input sentence. They are not effective to find nested NEs in a sentence. For example, ``Guizhou University'' is an organization NE, where ``Guizhou'' is  also a NE indicating the location of the university. In this case, outputting a label sequence cannot resolve the nested structure. Due to the reason that nested structures are effective to represent semantic relationships of entities (e.g., affiliation, ownership, hyponymy), they are widely used in natural languages. For example, in the GENIA corpus and ACE corpus, the nesting ratio is 35.27\% and 33.90\%, respectively~\cite{ohta2002genia,doddington2004automatic}.

Span classification is an effective method to recognize nested NEs. It generates NE spans from a sentence in a process known as region proposal, then outputs a class label for each possible NE span. It has two advantages to support nested NE recognition. First, nested NE structure can be resolved as separated NE spans. Second, the classification can be implemented on a span representation, which encode global semantic relevant to a predicated NE. At current, span classification has achieved a great attention. However, many span classification models enumerate all possible NE spans in a sentence, which suffer from a high computing complexity and the data unbalance problem. Therefore, many related work only verify possible NE spans upto a certain length (e.g., \cite{xu2016fofe,sohrab2018deep,xia2019multi}), or filter unlikely NE spans with predefined thresholds \cite{tan2020boundary} or NE boundary cues \cite{chen2019recognizing,zheng2019boundary}. For example, Chen et al. \cite{chen2019recognizing} verify NE spans combined from detected NE boundaries. Lin et al. \cite{lin2019sequence} apply a point network to recognize span boundaries relevant to the head word of a NE mention. The main problem for span classification is that, due to the reason of computation complexity and data imbalance, it is difficult to enumerate all NE candidates in a sentence. If a NE is not enumerated in the region proposal process, it cannot be recognized by span classification.

In our study, we found that there are some similarities between named entity recognition and object detection. For example, named entities in a sentence and objects in an image have similar spatial structures, e.g.,  {\em flattened}, {\em nested} and {\em discontinuous} \cite{uijlings2013selective}. The task to recognize them can be modelled as a span (or region) classification problem. The main difference between named entities and objects is that the former uses a discrete position representations. Because deep neural network has the ability to transform multimodal signals into an abstract semantic space \cite{zhang2019multimodal},  in this paper, we represent positions of NEs as continuous values. Then, a regression operation can be introduced to regress boundaries of NEs for locating NEs in a sentence the same way as detecting object detection in an image. 

In this paper, motivated by techniques developed in object detection of computer vision, we generate abstract NE representations from sentence in a region proposal process. The generated NE representations is named as textual bounding boxes (or bounding boxes in short). Every bounding box is annotated with three parameters to indicate its entity type, start position and length in a sentence. Then, instead of predicting the class of a NE candidate, a boundary regression operation is applied to refine its spatial location in a sentence. Based on boundary regression, we design a boundary regression (BR) model to support nested NE recognition. It is a multiobjective learning framework composed of a basic network module, a region proposal module, a classification module and a regression module. The  basic network transforms each sentence into abstraction representation. Then, a region proposal network is applied to generate bounding boxes. They are fed into the classification module fro class prediction and regression module for boundary regression. In the training process, in addition to maximizing the confidence scores of a NE, a linear layer is used to minimizes its location offset relative to a true NE.

The BR model simultaneously predicts the classification score of a NE candidate and  refine its spatial location in a sentence. The contributions of this paper include the following:
\begin{itemize}
\item[1)] Positions of NEs in a sentence are represented as continuous values to support NE boundary regression. It has the advantage to resolve nested NEs and locate NEs with boundary regression operation.

\item[2)] A bounding box based multiobjective learning model is designed to support nested NE recognition. It supports simultaneously predicts the class probability and refines spatial  locations of NEs in a sentence.
\end{itemize}

The structure of this paper is organized as follows. Before discussing the details of this model, our motivation is first discussed in Section \ref{sec:motivation}. Section \ref{sec:model} presents the definition of boundary regression and the architecture of the BR model. Experiments are conducted in Section \ref{sec:evaluation}, where several issues about the BR model are discussed. Section \ref{sec:related_work} introduce related work. Section \ref{sec:future_work} gives the conclusion of this paper.

\section{Motivation}
\label{sec:motivation}

The motivation of BR model is inspired by techniques developed in object detection in computer vision. From our intuition, a sentence is a one-dimensional linear textual stream, and an image is a two-dimensional pixel patch. They are totally different in external representations. However, in recent years, based on deep neural networks, language and vision can be embedded into a distributed representation and mapped into an abstract semantic space. Therefore, the combination of natural language processing and computer vision has become popular in research communities, e.g., the text retrieval approach in videos \cite{sivic2003video} and multimodal deep learning \cite{zhang2019multimodal}.

\begin{figure}[htbp]
	\centering
	\includegraphics[width=8cm]{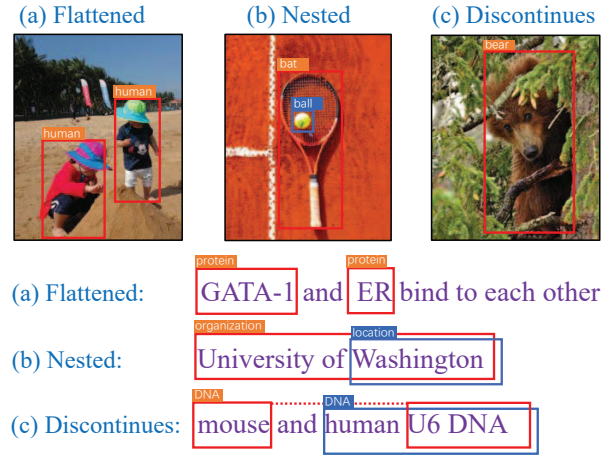}
	\caption{Similarity between Entities and Objects}
	\label{fig:correspondences}
\end{figure}

As shown in Figure \ref{fig:correspondences}, spatial patterns of entities in sentences and objects in images have similar structures.

The structures between objects (or entities) can be roughly divided into three categories: {\em flattened}, {\em nested} and {\em discontinuous} \cite{uijlings2013selective}. Flattened entities (or objects) are spatially separated from each other. In nested structure, two or more entities or objects are overlapped with each other. The discontinuous structure refers to disclosed objects (or entities). For example, ``HEL, KU812 and K562 cells'' contains three entities: ``HEL cells'', ``KU812 cells'' and ``K562 cells''. The first two NEs are discontinuous. Because the discontinuous structure can be transformed into nested structure \cite{chen2020boundary}. For example, the above examples can be processed as three nested NEs: ``HEL, KU812 and K562 cells'', ``KU812 and K562 cells'' and ``K562 cells''. In the paper, we only consider the flattened and nested structures. 

Object detection is a fundamental task in computer vision, which classifies regions of an image for predicting locations of objects. In the early stage,  the task is implemented in a multistage pipeline, where a region proposal step is applied to select coarse proposals. Another tendency for object detection adopts end-to-end architectures. A deep neural network is first adopted to map an input image into abstract representations known as {\em conv} feature maps. Then, proposals are generated from these feature maps. A proposal is an abstract representation of a possible object. Parameters are defined to indicate its location and shape in an image. Finally, a multiobjective learning framework is designed to simultaneously locate objects and predict the class probability.

\begin{figure}[h]
	\centering
	\includegraphics[width=7.5cm]{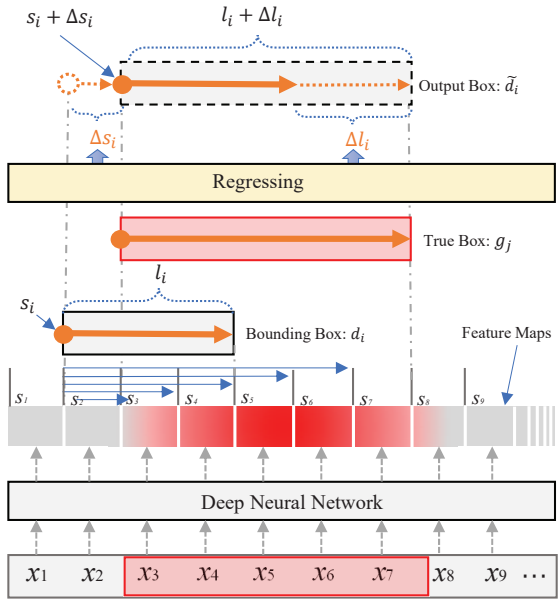}
	\caption{Boundary Regression}
	\label{fig:operation}
\end{figure}

Motivated by techniques developed in computer vision, we adopt a region proposal network to generate abstract NE representations (referred as bounding boxes). Every bounding box is annotated with three parameters to indicate its entity type, start position and length in a sentence.  Then, the spatial locations of NEs are represented as real values. It enables a regression operation for locating NEs in a sentence. The concept to regress boundaries for NE recognition is visualized in Figure \ref{fig:operation}.

As shown in Figure \ref{fig:operation}, an input sentence is first mapped into {\em recu} feature maps by a deep neural network. The feature maps can be seen as an abstract representation of an input sentence. Every feature map denotes a representation of an NE boundary, which can be bounded with others to generate bounding boxes (an example $d_i$ is shown in Figure \ref{fig:operation}). Every bounding box has two parameters denoting its position and shape in a sentence. If a bounding box correctly matches to a true NE, the box is referred to as a ``ground truth box'' (or truth box, e.g., $g_j$ in Figure \ref{fig:operation}).

Every box has two parameters to indicate its position ($s_i$) and shape (or length) ($l_i$)\footnote{In this paper, the position parameter ($s_i$) and shape parameter ($l_i$) of an NE are also referred to as location parameters.}. The regression operation respectively predicts the position offset and shape offset ($\Delta s_i$ and $\Delta l_i$) relative to a truth box. Finally, in the output, locations of the recognized NEs are updated as $\tilde{d}_i=\{s_i+\Delta s_i, l_i+\Delta l_i\}$. Because the outputs of a regression operation are continuous values, they are rounded to the nearest word boundary locations.

To design an end-to-end multiobjective learning architecture for boundary regression, we should carefully take the following four issues under consideration:

1) {\bf  representation}: object detection usually uses stacked convolutional layers to map an image into {\em conv} feature maps. In language processing, the recurrent (or attention) neural network is more effective in capturing the semantic dependency in a sentence. In this paper, the abstract representations of input sentences are referred to as {\em recu} feature maps.

2) {\bf region proposal}: in the {\em recu} feature map, a feature map position can be considered as an abstract representation of a possible entity boundary. It can be bounded with other feature maps to generate bounding boxes. Every bounding box is an abstract representation of an NE candidate labelled with its location information and class category.

3) {\bf multiobjective learning}: because the recurrent neural network can learn the semantic dependency, a bounding box contains semantic information about the whole sentence. In addition to predicting conditional class probabilities on a bounding box, a linear layer can be stacked to predict its location in a sentence.

4) {\bf maximum of overlapping neighbourhoods}: in the prediction process, every bounding box will approach a true bounding box. They are overlapped in the neighbourhood of a true bounding box. It is necessary to collect the most likely matched bounding boxes from overlapped bounding boxes.

According to the above discussion, we designed an end-to-end multiobjective learning architecture for boundary regression. The architecture of the BR model is given in the following section.


\section{Model}
\label{sec:model}

\begin{figure}[h]
	\centering
	\includegraphics[width=8.8cm]{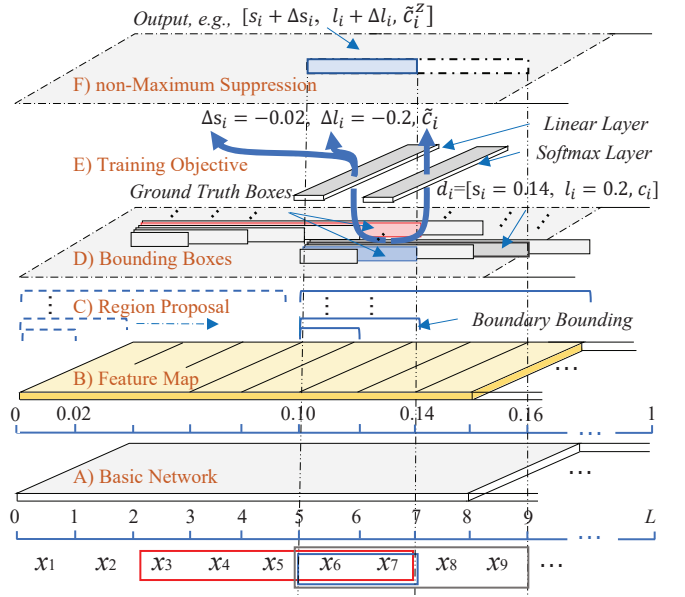}
	\caption{An input sentence is input into a {\em basic network}, which maps the sentence into a {\em recu feature map}. {\em Region proposal} generates {\em bounding boxes} from the {\em recu feature map}. Then, the model is trained to satisfy the {\em training objective}. {\em Non-maximum suppression} is adopted to produce a final decision.}
	\label{fig:regressing}
\end{figure}

In this paper, instead of modelling the NE recognition task as a classification problem, we frame the task as a multiobjective optimization process. In this framework, in addition to outputting discretized entity categories, a regression operation is integrated into a deep network for predicating the location offset of an NE candidate relative to a true NE in a sentence. The structure of the BR model is shown in Figure \ref{fig:regressing}.

As Figure \ref{fig:regressing} shows, six specific issues (referred to from A to F) are highlighted in the BR model. They are discussed as follows.

\subsection{Basic Network}
\label{sec:basic_network}

In a neural network model, the values of inputs represent the tense of signals. Therefore, words in a sentence are traditionally represented as high-dimensional one-hot vectors. At current, deep neural network has two advantages to support automatically extracting semantic features from raw inputs: word embedding and feature transformation. Word embedding is applied to map every word into a distributed representation, which encodes semantic information learned from external resources. In feature transformation, many types of neural layers (CNN \cite{c14}, LSTM \cite{hochreiter1997long} or Attention \cite{vaswani2017attention}) can be stacked to support designed feature transformation for capturing syntactic and semantic features of a sentence, which avoid the need for manually feature engineering.

For feature extraction, there are two differences between object detection and NE recognition. First, the detection of objects is mainly based on internal features of objects. Therefore, an object moving in an image exerts less influence on the object detection. However, entities have a strong semantic dependency in a sentence. Second, in nested objects, features in the bottom object are blocked by upper objects, which makes challenge for bottom object detection. While nested NEs share the same context in a sentence. It is important to lean dependent features relevant to considered NEs. Therefore, comparing with object detection, encoding semantic dependencies in a sentence is more important for NE recognition.  In deep architectures, a recurrent neural network or an attention network is helpful to capture the semantic dependency between words.

In our BR model, we adopted a basic network consisting of an embedding layer and a Bi-LSTM layer. The embedding layer is adopted to map a sentence into a distributed representation, where words (or characters) are embedded into vectors by a lookup table pretrained with unsupervised algorithms. Then, a Bi-LSTM is implemented to encode the semantic dependencies in a sentence. To simplify the region proposal step, we set the length of the input sentence as a fixed number, denoted as $L$. Longer or shorter sentences are trimmed or padded, respectively.

\subsection{Feature Map}
\label{sec:feature_map}

The output of the basic network is denoted as {\em recu} feature maps, where a feature map is an abstract token representation of the input. In the BR model, it also denotes as the feature map layer, which represent high-order abstract features of a sentence integrated with dependent semantics between words. In computer vision, images have an invariance property for a zooming operation. Object detection can benefit from multigranularity representations, where the region proposal can be implemented on multiscale feature maps to generate bounding boxes with different scales. In natural language processing, it is difficult to condense a textual sequence into multigranularity representations. At present, for each input sentence, we generate a single {\em recu} feature map layer, which can be seen as a high-order abstract representation of an input sentence. Each feature map position corresponds to a possible entity boundary. Because we adopt continuous location representation, the position of feature maps is normalized into the interval $[0,1]$ to support regression operation.

The feature map layer is mainly applied to support region proposal for generating abstract NE representations. Instead of directly generating NE candidates from a sentence (e.g., Sohrab et al.~\cite{sohrab2018deep}), generating abstract NEs from the feature map layer can share parameters in the bottom network. It reduces the computational complexity and enables more potent nonlinear function approximators to enhance model discriminability.

\subsection{Region Proposal}
\label{sec:proposals}

A feature map corresponds to an abstract representation of a possible NE boundary in a sentence. Each feature map can be set as a start position and combined with right feature maps to generate bounding boxes with different lengths. In this paper, for every feature map, we enumerate $K$ bounding boxes from left to right. The value $K$ is a predefined parameter indicating the longest NE candidate. It is similar to an exhaustive enumeration method, which verifies every possible NE candidate up to a certain length (e.g., Sohrab et al.~\cite{sohrab2018deep}). The difference is that bounding boxes are referred by their spatial locations in a sentence, which can be used to filter bounding boxes those are unlikely to be a truth box (discussed in Section \ref{sec:boxes}). It reduces the computational complexity and decreases the influence caused by negative examples.

To show the potential ability of the BR model to locate NEs those are missed in the region proposal process, in this experiment, we also propose an interval enumeration strategy. For every feature map, we enumerate bounding boxes from left to right with lengths [1, 3, 5, 7, 11, 15, 20]\footnote{We set the longest length of NEs to be 20. They cover 99.05\% of NEs in the Chinese ACE corpus.}.  In the training process, all ground truth boxes in training data are included to train the classifier. In the testing process, only bounding boxes with lengths [1, 3, 5, 7, 11, 15, 20] are verified. Comparing with exhaustively enumeration with lengths from 1 to 20, the interval enumeration reduces about 65\% computational overhead. For convenience, we refer to the BR model with interval enumeration as ``BR$_{\text{int}}$''.  The BR model implemented on exhaustive enumeration is referred to as ``BR$_{\text{exh}}$''. 

\subsection{Bounding Boxes}
\label{sec:boxes}

A bounding box is a high-order abstract representation of a possible NE generated from feature maps by region proposal. Because feature maps are transformed by a basic network which consists of a recurrent neural network or an attention network, each bounding box contains contextual features about a possible NE. Using class labels and location parameters of bonding boxes, a softmax layer and a linear layer can be set to predict their class probabilities and learn the location offset relative to a ground truth box. In the following, we give formal definitions about the bounding box.

Let $\mathbf{D}=\{d_1,d_2,\cdots,d_M\}$ denote a bounding box set generated from an input sentence $S$. $M$ is the size of $\mathbf{D}$.  Each bounding box $d_i \in \mathbf{D}$ has 3 parameters: $d_i^s$, $d_i^l$ and $\mathbf{c}_i$. Parameters $d_i^s$ and $d_i^l$ are two real numbers denoting the start position and length of $d_i$ in a sentence, respectively. The end position of $d_i$ can be computed as $d_i^s + d_i^l$. Parameter $\mathbf{c}_i=(c_i^1, c_i^2,\cdots,c_i^Z)$ $(c_i \in \{0,1\})$ is a one-hot vector representing the entity type of $d_i$, where $Z$ is the number of entity types. Therefore, a bounding box $d_i$ can be referred to as a three-tuple $d_i=\langle d_i^s, d_i^l, \mathbf{c}_i \rangle$. If a bounding box corresponds to a true NE, it is referred to as a ground truth box and represented as $g_j=\langle g_j^s, g_j^l, \mathbf{c}_j \rangle$.

Bounding boxes are labelled with location parameters. Borrowed from the intersection over union (IoU) developed in computer vision \cite{everingham2010pascal}, the overlapping ratio between two bounding boxes can be measured by the IoU value.

Let $d_i$ and $g_j$ be a bounding box and a ground truth box, respectively. The IoU value between them is computed as:
\begin{equation}
IoU(d_i,g_j)=\frac{span(d_i) \cap span(g_j)}{span(d_i) \cup span(g_j)}
\end{equation}
where the function $span(d_i)$ represents the range of a bounding box in feature maps. If a bounding box has a large IoU value, it is highly overlapped with a ground truth box. A high overlapping ratio indicates that a bounding box contains adequate contextual features about a true NE, which guarantees learning of the location offset relevant to a truth box. Otherwise, if the IoU value of a bounding box is lower than a predefined threshold, it denotes a false NE. It is used to train a classifier for identifying false NEs.

Let $\mathbf{D}_G$ represent the set of all ground truth boxes in $\mathbf{D}$. We define two sets as follows:
\begin{equation}
\label{equ:gamma}
\begin{split}
 & \mathbf{D}_p=\{d_i|d_i \in \mathbf{D}, \exists g_j\in \mathbf{D}_G(IoU(d_i,g_j)\geqslant \gamma) \}\\
 & \mathbf{D}_n= \{d_i| d_i\in \mathbf{D}, \forall g_j\in \mathbf{D}_G(IoU(d_i,g_j) < \gamma) \}
\end{split}
\end{equation}
where $\gamma$ is a predefined threshold. $\mathbf{D}_G$ is a subset of $\mathbf{D}_p$, where $\gamma$ is equal to 1.

In this paper, $\mathbf{D}_p$ is referred to as the positive bounding box set, and $\mathbf{D}_n$ is referred to as the negative bounding box set. In region proposal, a large number of negative bounding boxes will be generated, which lead to a significant data imbalance problem. This is also computationally expensive. In the training process, we collect $\mathbf{D}_p$ and $\mathbf{D}_n$ with a ratio of 1:3 for balancing positive and negative samples. This guarantees faster optimization and a stable training process.

Given a bounding box $d_i$, its relative ground truth box is identified as:
\begin{equation}
\label{equ:selected}
g_j=\mathop{\arg \max}_{g\in \mathbf{D}_G} IoU(d_i,g)
\end{equation}

Given a ground truth box $g_j$, all bounding boxes of $\mathbf{D}_p$ satisfying Equation \ref{equ:selected} are referred to as $\mathbf{D}_{g_j}$. They are the neighbourhoods of $g_j$. This is formalized as:
\begin{equation}
\label{equ:neighborhoods}
\mathbf{D}_{g_j}=\{ d_i | \text{$d_i$ is a neighbourhood of $g_j$.}\}
\end{equation}

It is important to know that, in the training data, all bounding boxes in $\mathbf{D}_{g_j}$ are labelled with a positive class tag the same as $g_j$. This labelling strategy is different from the traditional method in which, if the start and end boundaries of an NE candidate are not precisely matched to a true NE, it is labelled with a negative class tag. The reason for this will be discussed in detail in Section \ref{sec:loss}. For consistency, in this paper, we use the term ``positive box'' referring to a bounding box with an IoU value relevant to a ground truth box larger than $\gamma$. The term ``truth box'' refers to a bounding box, which has a location precisely matched to a real NE.
 
Based on $\mathbf{D}_G$ and Equation \ref{equ:selected}, $\mathbf{D}_p$ can be partitioned into a set $\mathcal{D}_p=\{\mathbf{D}_{g_1},\mathbf{D}_{g_2},\cdots\}$. Therefore, $\mathcal{D}_p$ is a partition of $\mathbf{D}_p$. If $i \neq j$, then $\mathbf{D}_{g_i} \cap \mathbf{D}_{g_j} = \emptyset$. Every bounding box $d_i \in \mathbf{D}_p$ belongs to a $\mathbf{D}_{g_j} \in \mathcal{D}_p$.  

For convenience, Table \ref{tab:appendix} lists the definitions about different bounding box sets. Their roles to support boundary regression will be discussed in the following subsection.
\begin{table}[!htbp]
\caption{Bounding Box Sets of Different Types}
\label{tab:appendix}
\begin{center}
\begin{tabular}{l|l}
\hline
\multicolumn{1}{c|}{\bf Symbol} & \multicolumn{1}{c}{\bf Meaning } \\ \hline
$\mathbf{D}$ &  A bounding box set generated from a sentence $S$;  \\ \hline
$\mathbf{D}_G$ & The set of all ground truth boxes in $\mathbf{D}$;  \\\hline
$\mathbf{D}_p$ & The positive bounding box set;  \\ \hline
$\mathbf{D}_n$ & The negative bounding box set;    \\ \hline
$\mathbf{D}_{g_j}$ & The neighbourhoods of $g_j$;   \\ \hline
$\mathcal{D}_p$ & A partition of $\mathbf{D}_p$;   \\ \hline
\end{tabular}
\end{center}
\end{table}

\subsection{Training Objective} 
\label{sec:loss}

Let $\hat{d}_{ij}^s = (g_j^s - d_i^s)/g_j^l$ and $\hat{d}_{ij}^l=log(g_j^l/d_i^l)$ be the normalized position offset and shape offset between $d_i$ and $g_j$. Given a bounding box $d_i$, the BR model outputs 3 parameters: $\Delta d_i^s$, $\Delta d_i^l$ and $\tilde{\mathbf{c}}_i$. $\Delta d_i^s$ and $\Delta d_i^l$ denote the predicted position offset and shape offset of $d_i$ relative to $g_j$. $\tilde{\mathbf{c}}_i$ is a confidence score that reflects the confidence that a box contains a true NE. As Figure \ref{fig:regressing} shows, $\Delta d_i^s$ and $\Delta d_i^l$ are regressed by a linear layer, while the classification confidence score $\tilde{\mathbf{c}}_i=(\tilde{c}_{i}^0, \tilde{c}_{i}^1,\cdots,\tilde{c}_{i}^Z)$  is predicted by a softmax layer.

For every $d_i \in \mathbf{D}_p$, the location offset of $d_i$ relates to a ground truth box that is predicted by a linear layer. A characteristic function $E_{ij}^z=\{0,1\}$ is defined to indicate that a default box $d_i$ is matched to a relative ground truth box $g_j$ selected by Equation \ref{equ:selected}. In the training process, the regression operation updates $\Delta d_i^s$ and $\Delta d_i^l$ for the purpose of approaching $\hat{d}_{ij}^s$ and $\hat{d}_{ij}^l$, respectively. The location loss can be computed as:
\begin{small}
\begin{equation}
\label{equ:loc_loss}
\! \!   L_{loc}(x, s, l)\! =\! \! \! \! \sum_{g_j \in \mathbf{D}_G} \! \! \frac{1}{N} \! \! \left( \! \sum_{d_i \in \mathbf{D}_{g_j}} \! \sum_{h \in \{s, l\}} \! \! \! \! E_{ij}^z \! \text{Smooth}_{L_1}(\Delta d_i^h \!  - \!  \hat{d}_{ij}^h) \! \right)
\end{equation}
\end{small}
where $N=|\mathbf{D}_{g_j}|$ is the cardinality of $\mathbf{D}_{g_j}$. It is used to normalize the weight between $g_j \in \mathbf{D}_G$. $\text{Smooth}_{L_1}$ is a robust $L_1$ loss that quantifies the dissimilarity between $d_i$ and $g_j$. It is less sensitive to outliers \cite{girshick2015fast}.

Equation \ref{equ:loc_loss} shows that only the positive bounding box set $\mathbf{D}_p$ is adopted to compute the location loss. For every ground truth box $g_j \in \mathbf{D}_G$, its neighbourhoods $\mathbf{D}_{g_j}$ are used to generate the location loss. This setting is natural because neighbourhoods contain sufficient contextual features about an NE to support boundary regression. On the other hand, negative bounding boxes are far from any ground truth box. Because the vanishing gradient problem, it is difficult to precisely regress their location offsets.

When minimizing the location loss, bounding boxes belonging to $\mathbf{D}_{g_j}$ will approach the ground truth box $g_j$. Therefore, all bounding boxes in $\mathbf{D}_{g_j}$ are given a class tag which is the same as ground truth box $g_j$.

Confidence loss is a softmax loss over multiple class confidences. It is given as follows:
\begin{equation}
L_{con}(x,c)= -\sum_{d_i \in \mathbf{D}_p} E_{ij}^z log(\tilde{c}_i^z) - \sum_{d_i \in \mathbf{D}_n} log(\tilde{c}_i^0)
\end{equation}
where $\tilde{c}_i^z=exp(\tilde{c}_i^z)/\sum_{z=0}^{Z}exp(\tilde{c}_i^z)$, $\tilde{c}_i^0$ is the confidence score indicating that an example is negative. A key issue about the classification is that the confidence score should be estimated based on NE representations with refined spatial locations in a sentence.

The total loss function combines the location loss and confidence loss:
\begin{equation}
L(x, s, l, c) =  L_{loc}(x, s, l) + \alpha L_{con}(x,c)
\end{equation}
where $\alpha$ is a predefined parameter balancing the weight between the location loss and confidence loss. The {\em training objective} is to reduce the total loss of the location offset and class prediction. In the training process, we optimize their locations to improve their matching degree and maximize their confidences.


\subsection{Non-Maximum Suppression}
\label{sec:suppression}

In the prediction process, the BR model outputs a set of bounding boxes for each input sentence, referred to as $\mathbf{D}=\{d_1,d_2, \cdots,d_M\}$. Every box $d_i\in \mathbf{D}$ has 3 outputs: $\Delta d_i^s$, $\Delta d_i^l$ and $\tilde{\mathbf{c}}_i$, respectively indicating the position offset, shape offset and class probability of $d_i$ relative to a truth box. After $\Delta d_i^s$ and $\Delta d_i^l$ are resized as $\Delta s_i$ and $\Delta l_i$, an predicated NE can be located as: [$s_i+\Delta s_i$, $l_i+\Delta l_i$, $\tilde{c_i}$]\footnote{An example of the output is shown in Figure \ref{fig:regressing}.}.

The output $\mathbf{D}$ contains a large number of boxes, but many of them are overlapped. Non-maximum suppression (NMS) is implemented in the prediction process to produce the final decision, which selects truth boxes from overlapped neighbourhoods. The NMS algorithm is shown in Table \ref{tab:NMS}.

\begin{table}[htbp]
\begin{center}
\small
\caption{The NMS algorithm of the BR model}
\label{tab:NMS}
\begin{tabular}{p{8cm}}
\hline
Input: $\mathbf{D}=\{d_1,d_2,\cdots,d_M\}$, a threshold $\lambda$.\\
Output: $\mathbf{D}_e$ recognized NEs. \\\hline
1: Sort $\mathbf{D}$ according confidence scores in descending order;\\
2: While (if $\mathbf{D}$ is not empty)\{\\
3: \quad \quad Select $d=d_1$ from $\mathbf{D}$;\\
4: \quad \quad Delete $d_1$ from $\mathbf{D}$;\\
5: \quad \quad For $\forall d_i \in \mathbf{D}$, if $IoU(d,d_i)>\lambda$, delete $d_i$ from $\mathbf{D}$;\\
6: \quad \quad Add $d$ to $\mathbf{D}_e$; \ \}\\
\hline
\end{tabular}
\end{center}
\end{table}

It is a one-dimensional NMS algorithm that selects nested NEs from overlapped positive boxes. The NMS algorithm searches local maximized elements from overlapping neighbourhoods in which a smaller number of high-confidence boxes are collected. The threshold is adopted to control the overlapping ratio between neighbourhoods. In our experiments, the value of $\lambda$ is set as 0.6.

\section{Experiments}
\label{sec:evaluation}


In our experiments, the ACE 2005 corpus \cite{doddington2004automatic} and the GENIA corpus \cite{kim2003genia} are adopted to evaluate the BR model. In order to show the performance of the BR model to recognize flattened NE structure, in Section \ref{sec:ablation}, the BR model is also evaluated on the OntoNotes 5.0 \cite{pradhan2013towards} and CoNLL 2003 English \cite{sang2003introduction} corpora.

The ACE 2005 corpus is collected from broadcasts, newswires and weblogs. It is the most popular source of evaluation data for NE recognition. The corpus contains three datasets: Chinese, English and Arabic. In this paper, the BR model is mainly evaluated on the ACE Chinese corpus. To show the extensibility of the BR model regarding other languages, it is also evaluated on the ACE English corpus and the GENIA corpus.

The GENIA corpus was collected from biomedical literature. It contains 2,000 abstracts in MEDLINE by PubMed based on three medical subject heading terms: human, blood cells and transcription factors. This dataset contains 36 fine-grained entity categories. In the GENIA corpus, many NEs have discontinuous structure. They are transformed into nested structures by holding the discontinuous NE as a single mention.

\begin{figure}[htbp]
	\centering
	\includegraphics[width=8.5cm]{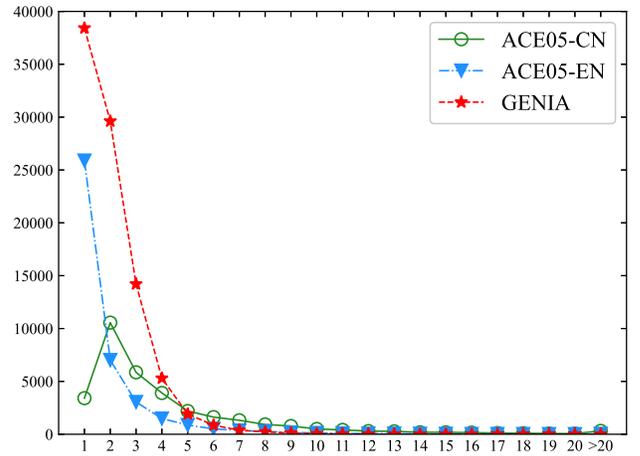}
	\caption{Distributions of NE lengths}
	\label{fig:lengths}
\end{figure}

In the ACE Chinese dataset, there are 33,238 NEs in total. The number of NEs in the ACE English dataset is 40,122. The GENIA corpus is annotated with 91,125 NEs. The distributions of NE lengths in the three corpora are shown in Figure \ref{fig:lengths}.

In the basic network, the default length of sentence $L$ is 50. Sentences with longer or shorter length are trimmed or padded, respectively. In the total loss function, $\alpha=1$ is used. Two BERT$_{BASE}$ \cite{devlin2018bert} model are tuned by implementing the innermost and outermost NE recognition tasks, respectively. Then, every sentence is encoded into two sequence of vectors by two tuned BERT models, where every word in a sentence is encoded as a concatenated $2\times 768$ dimensional vector. It is fed into a Bi-LSTM layer, which outputs a $2\times 128$ dimensional {\em recu} feature map. In the training process, word representations are fixed and not subject to further tuning.

In region proposal, two strategy have been introduced: exhaustive enumeration and interval enumeration, which corresponds to two BR models referred to as ``BR$_{\text{exh}}$'' and ``BR$_{\text{int}}$''.  The BR$_{\text{exh}}$ model exhaustively enumerates all NEs with length up to 6. We ignore NEs with length larger than 6. In the BR$_{\text{int}}$ model, we intermittently enumerate bounding boxes from left to right with lengths [1, 3, 5, 7, 11, 15, 20]. To collect the positive bounding box set $\mathbf{D}_p$ to train the linear layer,  $\gamma$ is set as 0.7 and 0.6 for BR$_{\text{exh}}$ and  BR$_{\text{int}}$, respectively. The quantitative test to set $\gamma$ is discussed in Section \ref{tab:Influence_of_IoU}.

In the output layer, a correct NE requires that start and end boundaries of an NE are precisely identified. Because the BR model uses a regression operation to predicate spatial locations of NEs in a sentence, all entity locations are mapped into interval $[0,1]$ for a smooth learning gradient. Therefore, the output of BR is rounded to the nearest character location. 

\subsection{Comparison with related work}

To show the superiority of our model, we first compare the BR model with related work. It is first evaluated on the Chinese corpus. Then, to show the extensibility of the BR model, the BR model is transformed to the English corpus for further assessment.

\vspace{0.3cm}
\subsubsection{Evaluation in the Chinese Corpus} \
\vspace{0.1cm}

In the Chinese corpus, we first conduct a popular sequence model (Bi-LSTM-CRF) \cite{huang2015bidirectional}. It consists of an embedding layer, a Bi-LSTM layer, an MLP layer and a CRF layer. The embedding layer and Bi-LSTM layer have the same settings as the basic network of the BR model. We adopt cascading and layering strategies to solve the nesting problem \cite{alex2007recognising}. The layering model proposed by Lin et al. \cite{lin2019sequence} is adopted for comparison.

In the Chinese ACE corpus, BA is a pipeline framework for nested NE recognition that has achieved state-of-the-art performance \cite{chen2019recognizing}. The original BA is a ``Shallow'' model, which uses a CRF model to identify NE boundaries and a maximum entropy model to classify NE candidates. NNBA is a neural network version, where the LSTM-CRF model is adopted to identify NE boundaries, and a multi-LSTM model is adopted to filter NE candidates.

In this experiment, the ``Adam'' optimizer is adopted. The learning rate, weight decay rate and batch size are set as 0.00005, 0.01 and 30, respectively. Shallow models refer to CRF-based models. In the BR model, we use the same settings as Chen et al. \cite{chen2019recognizing} to configure the basic neural network, where the BERT is adopted to initialize word embeddings. These models are implemented with the same data and settings as Chen et al. \cite{chen2019recognizing}. The results are shown in Table \ref{tab:comparing_chinese_model}. 

\begin{table}[!htbp]
\footnotesize
\begin{center}
\caption{Evaluation in the Chinese Corpus}
\label{tab:comparing_chinese_model}
\begin{tabular}{c|l|ccc}
\hline
 \multicolumn{2}{c|}{Model} & P(\%)& R(\%)& \multicolumn{1}{c}{ F(\%)}   \\\hline
\multirow{5}*{Shallow Models}& Innermost & 73.60 & 45.50 & 56.32  \\
 & Outermost     					& 72.60 & 45.54 & 55.97  \\
 & Cascading 						& 76.52 & 51.80 & 61.80 \\
 & Layering 						& 71.93 & 56.57 & 63.33  \\
 & BA \cite{chen2015boundary}       & 73.98 & 62.16 & 67.56   \\\hline
\multirow{5}*{Deep Models} & Innermost & 82.00 & 70.70 & 75.93  \\
 & Outermost     & 80.45 & 69.08 & 74.33  \\
 & Cascading  & 76.96 & 71.39 & 74.07 \\
 & Layering \cite{lin2019sequence} & 78.85 & 81.34 & 80.07  \\
 & NNBA \cite{chen2019recognizing} & 80.49 & 79.46 & 79.97  \\ \hline
 \multicolumn{2}{c|}{BR$_{\text{int}}$} & 72.35 & 73.71 & 73.02 \\
 \multicolumn{2}{c|}{BR$_{\text{exh}}$} & 85.95 & 84.39 & 85.16 \\
 \hline
\end{tabular}
\end{center}
\end{table}

\begin{table*}[htbp]
\footnotesize
\begin{center}
\caption{Evaluation in the English Corpus}
\label{tab:comparing_english_model}
\begin{tabular}{l|l|p{0.55cm}<{\centering}p{0.55cm}<{\centering}p{0.55cm}<{\centering}|p{0.55cm}<{\centering}p{0.55cm}<{\centering}p{0.55cm}<{\centering}}\cline{2-8}
 & \multirow{2}{*}{ Models } & \multicolumn{3}{c|}{GENIA} & \multicolumn{3}{c}{ACE}    \\\cline{3-8}
 			&  & P(\%) & R(\%) & F(\%) & P(\%) & R(\%) & F(\%)  \\\hline
Lu et al. \cite{lu2015joint}         	& Mention Hypergraphs & 72.5  & 65.2 & 68.7 & 66.3 & 59.2 & 62.5 \\
Katiyar et al. \cite{katiyar2018nested}	& Neural Hypergraph   & 76.7 & 71.1 & 73.8 & 70.6 & 70.4 & 70.5  \\
Ju et al. \cite{ju2018neural}         	& Layered-BiLSTM-CRF  & 78.5  & 71.3 & 74.7 & 74.2 & 70.3 & 72.2  \\
Wang et al. \cite{wang2018neural}     	& Stack-LSTM 		  & 77.0  & 73.3 & 75.1 & 76.8 & 72.3 & 74.5  \\
Lin et al. \cite{lin2019sequence}     	& Sequence-to-nuggets & 75.8  & 73.9 & 74.8 & 76.2 & 73.6 & 74.9  \\
Xia et al. \cite{xia2019multi}         	& MGEPN				  & - & - &- & 79.0 & 77.3 & 78.2  \\
Fisher et al. \cite{fisher2019merge}   	& BERT+Merge\&Label   & - & - &- & 82.7 & 82.1 & 82.4\\
Shibuya et al. \cite{shibuya2019nested}	& BERT+FLAIR		  & 76.3  & 74.7 & 75.5 & 85.94 & 85.69 & 85.82 \\
Strakova et al. \cite{strakova2019neural}	& BERT+Seq2Seq	  & -     & - & 78.3 &  - & - & 84.33  \\
Wang et al. \cite{wang2020pyramid} 		& BERT+Pyramid	      & 79.45 & 78.94 & 79.19  & 85.30 & 87.40 & 86.34 \\ 
Tan et al. \cite{tan2020boundary}   		&BERT+Boundary & 79.2 & 77.4 & 78.3 & 83.8 & 83.9 & 83.9 \\
Li et al. \cite{li2020unified} 		& BERT+MRC				  & 85.18 & 81.12 & 83.75 & 87.16 & 86.59 & 86.88  \\\hline 
\multirow{2}*{ Ours} &  \multicolumn{1}{c|}{BR$_{\text{int}}$} 			  & 80.81& 77.43 & 79.09  & 86.88  & 84.83& 85.84 \\
     &  \multicolumn{1}{c|}{BR$_{\text{exh}}$} 			  & 81.74 & 81.75 & 81.75  & 89.10  & 87.52 & 88.30  \\\hline
\end{tabular}
\end{center}
\end{table*}

In Table \ref{tab:comparing_chinese_model}, all deep models outperform shallow models because neural networks can effectively utilize external resources by using a pretrained lookup table and have the advantage of learning abstract features from raw input. In deep models, the performances of innermost and outermost models are heavily influenced by a lower recall rate, which is caused by ignoring nested NEs. The deep cascading model also suffers from poor performance because predicting every entity type by an independent classifier can not make full use of annotated data. The deep layering model is impressive. This model is produced by implementing two independent classifiers that separately recognize the innermost and outermost NEs. It offers higher performance, even outperforming the NNBA model. The reason for the improvement is that, in our experiments, entities with length exceeding 6 are ignored, which decreases the nesting ratio. Most of the nested NEs have two layers, which can be handled appropriately by the layering model. In Table \ref{tab:comparing_chinese_model}, the BR model exhibits the best performance.


The Chinese language is hieroglyphic. It contains very little morphological information (e.g., capitalization) to indicate word usage. Because there is a lack of delimitation between words, it is difficult to distinguish monosyllabic words from monosyllabic morphemes. However, Chinese has two distinctive characteristics. First, Chinese characters are shaped similar squares. They are known as square-shaped characters. Their locations are uniform. Second, because the meaning of a Chinese word is usually derived from the characters it contains, every character is informative. Therefore, character representations can effectively capture the syntactic and semantic information of a sentence. The BR model works well on the Chinese corpus.

\vspace{0.3cm}
\subsubsection{Evaluation in the English Corpus} \
\vspace{0.1cm}

In  the ACE English corpus and the GENIA corpus, we adopt the same settings as Lu et al. \cite{lu2015joint} to evaluate the BR model, where the evaluation data are divided according to the proportion 8:1:1 for training, developing and testing. In the GENIA corpus, researchers often report the performance with respect to five NE types (DNA, RNA, protein, cell line and cell type). To compare with existing methods, we generate results for the five NE types. 

In Table \ref{tab:comparing_english_model}, Lu et al. \cite{lu2015joint} and Katiyar et al. \cite{katiyar2018nested} represent nested NEs as mentioned hypergraphs. Ju et al. \cite{ju2018neural} feed the output of a BiLSTM-CRF model to another BiLSTM-CRF model. This strategy generates layered labelling sequences. The stack-LSTM \cite{wang2018neural} uses a forest structure to model nested NEs. Then, a stack-LSTM is implemented to output a set of nested NEs. Sequence-to-nuggets \cite{lin2019sequence} first identifies whether a word is an anchor word of an NE with specific types. Then, a region recognizer is implemented to recognize the range of the NE relative to the anchor word. MGEPN \cite{xia2019multi} and Merge\&Label are pipeline frameworks. They first generate NE candidates. Then, all candidates are further assessed by a classifier. FLAIR \cite{shibuya2019nested} extracts entities iteratively from outermost to innermost. Strakova et al. \cite{strakova2019neural} encode an input sentence into a vector representation. Then, a label sequence is directly generated from the sentence representation. Wang et al. \cite{wang2020pyramid} use a CNN to condense a sentence into a stacked hidden representation with a pyramid shape, where a layer represents NE candidate representations with different granularity. These models are all nesting-oriented models. Their performances are listed in Table \ref{tab:comparing_english_model}.

Table \ref{tab:comparing_english_model} shows that the performance of the GENIA corpus is lower than that of the ACE corpus. There are three reasons for this phenomenon. First, the GENIA corpus was annotated with discontinuous NEs. For the example mentioned in Section \ref{sec:motivation}, ``HEL, KU812 and K562 cells'' contains two  discontinuous NEs. Second, in the GENIA corpus, nested NEs may occur in a single word. For example, ``TCR-ligand'' is annotated as an ``other\_name'' entity, where it is nested with a ``TCR'' protein. Third, a large number of abbreviations are annotated in the GENIA corpus, which brings about a serious feature sparsity problem. Therefore, the performance is lower in the GENIA corpus.

In related work, many models also exhaustively verify every possible NE candidate with length up to 6, e.g., Xu et al. \cite{xu2016fofe}, Sohrab et al. \cite{sohrab2018deep}, Xia \cite{xia2019multi} and Tan et al. \cite{tan2020boundary}. Because limiting the length of NEs can reduce the influence caused by negative instances, these models achieve higher performance. In comparison with them, the BR$_{\text{int}}$ model significantly improves the performance. In the testing data, the ratios of NEs with lengths [1, 3, 5, 7, 11, 15, 20] in the ACE English and the GENIA corpora are 79.47\% and 58.34\%, respectively (the ratio in the ACE Chinese corpus is 39.89\%). Therefore, the BR$_{\text{int}}$ model also achieves competitive performance in the ACE English and the GENIA corpora. 

In Table \ref{tab:comparing_english_model}, all neural network-based models exhibit higher performance. Especially in the BERT models, the performance is improved considerably. Li et al. \cite{li2020unified} present a model based on machine reading comprehension, where manually designed questions are required to encode NE representations. It achieves higher performance with respect to the GENIA corpus. However, because this model benefits from prior knowledge and experience, which essentially introduce descriptive information about the categories, it is rarely used for comparison with related work. In comparison with related work in the English corpus, the BR model also shows competitive performance. 


\subsection{Ablation Study}
\label{sec:ablation}

In natural language processing, continuous location representation has not been used denoting to positions of linguistic units in a sentence. Therefore, the regression operation is rarely used to support information extraction. As we known, the BR model is the first attempt to locate linguistic units in a sentence by regression operation. To analyse the mechanism of boundary regression for nested NE recognition, we design a traditional NE classification model named as bounding box classifier (BBC) for comparison. It is generated by omitting the linear layer from the deep architecture in Figure \ref{fig:regressing}. In the output, only a softmax layer is adopted to predict the class probability for every bounding box.

\begin{table*}[htbp]
\footnotesize
\begin{center}
\caption{Feasibility of Boundary Regression}
\label{tab:influence_regression}
\begin{tabular}{c|r|p{0.55cm}<{\centering}p{0.55cm}<{\centering}p{0.55cm}<{\centering}||p{0.55cm}<{\centering}p{0.55cm}<{\centering}p{0.55cm}<{\centering}||p{0.55cm}<{\centering}p{0.55cm}<{\centering}p{0.55cm}<{\centering}}
\cline{3-11}
 \multicolumn{1}{c}{}&  \multicolumn{1}{c|}{}& \multicolumn{3}{c||}{ BBC (0.7) Model} & \multicolumn{3}{c||}{ BBC (1.0) Model } & \multicolumn{3}{c}{ BR$_{\text{exh}}$ Model}\\\hline
 TYPE & Number & P(\%) & R(\%) & \multicolumn{1}{c||}{ F(\%)} & P(\%) & R(\%) & \multicolumn{1}{c||}{ F(\%)} & P(\%) & R(\%) &\multicolumn{1}{c}{ F(\%)}   \\\hline
 VEH  & 499  & 46.15 & 71.32 & 56.04 	& 84.87 & 70.62 & 77.09 		& 79.36 & 69.93 & 74.34	 \\
 LOC  & 1,277    & 33.33 & 53.09 & 40.95 	& 76.11 & 49.81 & 60.21 	& 74.88 & 57.45 & 65.02 \\
 WEA  & 324      & 44.44 & 63.76 & 52.38 	& 78.57 & 63.76 & 70.40 	& 77.58 & 65.21 & 70.86 \\
 GPE   & 8,071   & 49.83 & 86.90 & 63.34 	& 87.29 & 85.72 & 86.50		& 86.24 & 86.77 & 86.50 \\
 PER   & 11,351& 46.21 & 90.65 & 61.22 	& 90.79 & 90.16 & 90.48		& 89.81 & 89.96 & 89.88   \\
 ORG   & 4,837   & 34.68 & 80.81 & 48.53 	& 82.56 & 79.46 & 80.98		& 82.79 & 81.19 & 81.98 \\
 FAC   & 1,194    & 34.56 & 67.33 & 45.67 	& 75.22 & 65.33 & 69.93		& 73.25 & 70.91 & 72.06\\\hline
 Total & 27,553   & 43.62 & 84.30 & 57.49 	& 87.00 & 83.27 & 85.10		& 85.95 & 84.39 & 85.16 \\\hline
\end{tabular}
\end{center}
\end{table*}

In this section, three experiments are conducted to show the usefulness of the regression operation. We first conduct two ablation studies to show the feasibility of boundary regression. In the first experiment, exhaustive enumeration is adopted in region proposal. The BBC model and the BR model are compared to show the ability of BR to refine spatial locations of NEs in a sentence. In the second experiment, the BR model is implemented on intermittently enumerated bounding boxes. The experiment shows the ability of the BR model to locate true NEs from mismatched NE candidates. The BR model is mainly designed to supported nested NE recognition. It also can be used to recognize flattened NEs. Therefore, in the third experiment, we evaluate the BR model on flattened NE recognition. The first experiment and the second experiment are conducted on the ACE Chinese corpus. The third experiment is implemented on two English corpora with flattened NE annotation: the OntoNotes 5.0 \cite{pradhan2013towards} corpus and the CoNLL 2003 \cite{sang2003introduction} corpus.

\vspace{0.3cm}
\subsubsection{Performance with exhaustive enumeration} \
\vspace{0.1cm}

In this experiment, we compare the the BR$_{\text{exh}}$ model with two BBC model: BBC (0.7) and BBC (1.0). The BBC (0.7) model is implemented on the same evaluation data as the BR model with $\gamma=0.7$  to collect positive bounding boxes. In the BBC (1.0) model, $\gamma=1.0$ is applied. Under this setting, the positive bounding box set and the negative bounding box set can be denoted as: $\mathbf{D}_G$ and $(\mathbf{D}_p \cup \mathbf{D}_n)-\mathbf{D}_G$. It means that every positive bounding box is precisely matched to a true ground truth box. Therefore, the BBC (1.0) model is a traditional classifier implemented on precisely annotated evaluation data. 

We implement the BBC model and the BR$_{\text{exh}}$ model with the same data and settings. The result is shown in Table \ref{tab:influence_regression}, where Column ``Number'' refers to the number of annotated NEs in the corpus. the performance is reported with respect to 7 true entity types. The ``Total" denotes to the micro-average on all entity types.

In NE recognition, a correct output requires that both the start and end boundaries are precisely matched to a manually annotated NE. Because the BBC model is a traditional classifier which cannot regress mismatched boundaries, as Table \ref{tab:influence_regression} shows, it suffers from significantly diminished precision caused by mismatched NE boundaries. The BBC (1.0) model is implemented on the evaluation data with $\gamma=1.0$, where boundaries of positive bounding boxes are precisely matched to true NEs. The result in Table \ref{tab:influence_regression} shows that, in comparison with the BBC (0.7) model, BBC (1.0) achieves higher performance.

In the BR$_{\text{exh}}$ model, because bounding boxes in $\mathbf{D}_p$ have a high overlapping ratio relevant to a ground truth box, they have sufficient semantic features with respect to a true NE for supporting boundary regression. In the prediction process, mismatched boundaries of bounding boxes can approach a ground truth box through the regression operation. In comparison with the BBC (0.7) model, mismatched boundaries can be revised, which considerably improve the performance. The result indicates that the regression operation really regresses boundaries and locates NEs in a sentence.

Comparing the BR$_{\text{exh}}$ model with the BBC (1.0) model, in both the BR model and the BBC (1.0) model, all NEs with length up to 6 have been enumerated and verified. In this condition, in the prediction process of the BR$_{\text{exh}}$ model, approaching to an NE that is already verified is less helpful to improve the performance. However, because the BR$_{\text{exh}}$ model can refine spatial locations of bounding boxes in $\mathbf{D}_p$ and share model parameters in the bottom network, a higher recall ratio can be achieved in the BR$_{\text{exh}}$ model, which improve the final performance.

\vspace{0.3cm}
\subsubsection{Performance of interval enumeration} \
\vspace{0.1cm}

In the second experiment, the BBC (1.0) model is compared with the BR$_{\text{int}}$, which only verifies bounding boxes with lengths [1, 3, 5, 7, 11, 15, 20] in the testing dataset. The results are listed in Table \ref{tab:superiority}.

\begin{table}[htbp]
\footnotesize
\begin{center}
\caption{Superiority of Boundary Regression}
\label{tab:superiority}
\begin{tabular}{c|ccc||ccc}
\cline{2-7}
 \multicolumn{1}{c|}{}& \multicolumn{3}{c||}{ BBC (1.0) Model } & \multicolumn{3}{c}{ BR$_{\text{int}}$ Model }\\\hline
 TYPE & P(\%) & R(\%) & \multicolumn{1}{c||}{ F(\%)} & P(\%) & R(\%) & \multicolumn{1}{c}{ F(\%)}    \\\hline
 VEH   & 86.53 & 25.42 & 39.30 	& 60.60 & 56.49 & 58.47 \\
 LOC   & 66.66 & 11.65 & 19.84 	& 53.99 & 43.55 & 48.21	 \\
 WEA   & 68.96 & 27.39 & 39.21  & 71.66 & 58.90 & 64.66	 \\
 GPE   & 82.34 & 25.17 & 38.56 & 82.65 & 78.87 & 80.72	 	\\
 PER   & 91.68 & 44.70 & 60.10  & 80.24 & 79.82 & 80.03	 \\
 ORG  & 82.10 & 32.01 & 46.06   & 70.09 & 65.09 & 67.50	 \\
 FAC   & 77.27 & 20.98 & 33.00  & 61.37 & 54.93 & 57.98	 \\\hline
 Total  & 86.67 & 34.00 & 48.84 & 76.46 & 73.00 & 74.69	\\\hline
\end{tabular}
\end{center}
\end{table}

Because the BBC is a traditional classifier which only assigns a class tag to every NE candidate, it cannot regress NE boundaries for locating possible NEs. Therefore, if a true NE is not enumerated in the testing data, it will be missed by the traditional classifier, which leads to greatly reduced recall. For example, in the ACE Chinese corpus, a sentence ``\begin{CJK}{UTF8}{gbsn}中国要把广西发展为连接西部地区和东南亚的桥梁\end{CJK}'' (China wants to develop Guangxi into a bridge connecting the western region and Southeast Asia) contains five NEs: ``\begin{CJK}{UTF8}{gbsn}中国\end{CJK}'' (China), ``\begin{CJK}{UTF8}{gbsn}广西\end{CJK}'' (Guangxi), ``\begin{CJK}{UTF8}{gbsn}西部地区\end{CJK}'' (the western region), ``\begin{CJK}{UTF8}{gbsn}西部\end{CJK}'' (the western), ``\begin{CJK}{UTF8}{gbsn}东南亚\end{CJK}'' (Southeast Asia), which correspond to five ground truth boxes: [0, 2, GPE]\footnote{In this triple, the parameters represent the start position, the length and the type of an NE in a sentence.}, [4, 2, GPE], [11, 4, LOC], [11, 2, LOC], and [16, 2, GPE]. In the BBC model, only ``\begin{CJK}{UTF8}{gbsn}东南\end{CJK}'' can be enumerated and verified, which considerably worsens the performance.

In the BR$_{\text{int}}$ model, suppose a true NE is missed in the region proposal process, if it is overlapped with one or more bounding boxes, the regression operation can refine their spatial locations in a sentence, which enables these boxes approaching the missing true NE. As in the previous example, in the BR$_{\text{int}}$ model, ``\begin{CJK}{UTF8}{gbsn}西部地区\end{CJK}'' cannot be enumerated in the testing data, but it is overlapped by at least two bounding boxes: [11, 3, ?] (``\begin{CJK}{UTF8}{gbsn}西部地\end{CJK}'') and [11, 5, ?] (``\begin{CJK}{UTF8}{gbsn}西部地区和\end{CJK}''), where ``?'' means that the class tag is unknown. Because their IoU values with the truth box [11, 4, LOC] is larger than 0.7 ( $\gamma>0.7$). They contain semantic information about ``\begin{CJK}{UTF8}{gbsn}西部地区\end{CJK}''. The softmax layer outputs a high confidence score on ``LOC''. More than anything, the offsets which are relevant to the truth box [11, 4, LOC] are learned, which enables the NE ``\begin{CJK}{UTF8}{gbsn}西部地区\end{CJK}'' to be recognized correctly.

\vspace{0.3cm}
\subsubsection{Evaluation in the Flattened Corpus} \
\vspace{0.1cm}

In this section, the OntoNotes 5.0 \cite{pradhan2013towards} and CoNLL 2003 English \cite{sang2003introduction} corpora are employed to evaluate the performance of BR model to recognize NEs with flattened structure. The OntoNotes corpus is collected from a wide variety of sources, e.g., magazine, telephone conversation, newswire, etc. It contains 76,714 sentences and annotated with 18 entity types. The CoNLL corpus consists of 22,137 sentences collected from Reuters newswire articles. It is split into 14,987, 3,466 and 3,684 sentences for training, developing and testing.

The BR model is compared with several SOTA models conducted on the OntoNotes and CoNLL corpora. Ma et al. \cite{ma2016end} is a BiLSTM-CNNs-CRF model, which automatically encodes semantic features from words and characters. Ghaddar et al. \cite{ghaddar2018robust} is also a BiLSTM-CRF model learning lexical features from word and entity type representations. Devlin et al. \cite{devlin2018bert} is the BERT framework, which is effective to learn semantic features from external resources. Li et al. \cite{li2020unified} is a model based on machine reading comprehension. Yu et al. \cite{yu2020named} uses a biaffine model to encode dependency trees of sentences. Luo et al. \cite{luo2020hierarchical} is also a BiLSTM model based on hierarchical contextualized representations. The result is shown in Table \ref{tab:Flattened}.

\begin{table}[htbp]
\footnotesize
\begin{center}
\caption{Evaluation in the Flattened Corpus}
\label{tab:Flattened}
\begin{tabular}{l|p{0.6cm}<{\centering}p{0.6cm}<{\centering}p{0.6cm}<{\centering}||p{0.6cm}<{\centering}p{0.6cm}<{\centering}p{0.6cm}<{\centering}}
\cline{2-7}
 \multicolumn{1}{c|}{}& \multicolumn{3}{c||}{ OntoNotes} & \multicolumn{3}{c}{ CoNLL }\\\hline
 Architecture & P(\%) & R(\%) & \multicolumn{1}{c||}{ F(\%)} & P(\%) & R(\%) & \multicolumn{1}{c}{ F(\%)}    \\\hline
 BiLSTM \cite{ma2016end}   & 86.04 & 86.53 &  86.28 & -  & - & 91.03	 	\\
 BiLSTM \cite{ghaddar2018robust}  & - & - & 87.95  & - & - & 91.37	 \\
BERT \cite{devlin2018bert}  & 90.01 & 88.35 & 89.16  & - & - & 92.8	 \\
MRC \cite{li2020unified}   & 92.98 & 89.95 & 91.11   & 92.33 & 94.61 & 93.04	 \\
Biaffine \cite{yu2020named}  & 91.1 & 91.5 & 91.3 & 93.7 & 93.3 & 93.5 	 \\
BiLSTM \cite{luo2020hierarchical}   & - & - & 90.37 	& - & - & 93.37	 \\\hline
\ \ \ BR$_{\text{int}}$   & 89.36 & 89.87& 89.61 & 91.32 & 92.99 & 92.15 \\
\ \ \ BR$_{\text{exh}}$  & 90.94  &  88.81 & 89.86 & 92.89  & 91.86 & 	92.37\\\hline
\end{tabular}
\end{center}
\end{table}

In Table \ref{tab:Flattened}, the compared models are all sequence models. There are three of them directly based on the BiLSTM network. Another tree models (the BERT, MRC and Biaffine) also applied BiLSTM as an inner structure for capturing semantic dependencies in a sentence. Because sequence models output a maximized labelling sequence for each input sentence, they are effective to encode syntactic and semantic structures in a sentence. Therefore, in flattened NE recognition, sequence models achieve the best performance.

Comparing the BR model with sequence models, the BR model can be seen as a span classification model, which applied a regression operation to refine spatial locations of NEs in a sentence. Because the classification is based on enumerated spans, due to the reason of the vanishing gradient problem, it is weak to encode long-distance semantic dependencies in a sentence for flattened NEs. Even so, as Table \ref{tab:Flattened} shows, the BR model also achieves competitive performance in flattened NE recognition.

\begin{figure*}[h]
  \centering 
  \subfigure[IoU ($\gamma$) of BR$_{\text{exh}}$]{\label{fig:IoU_for_6} \includegraphics[width=3in]{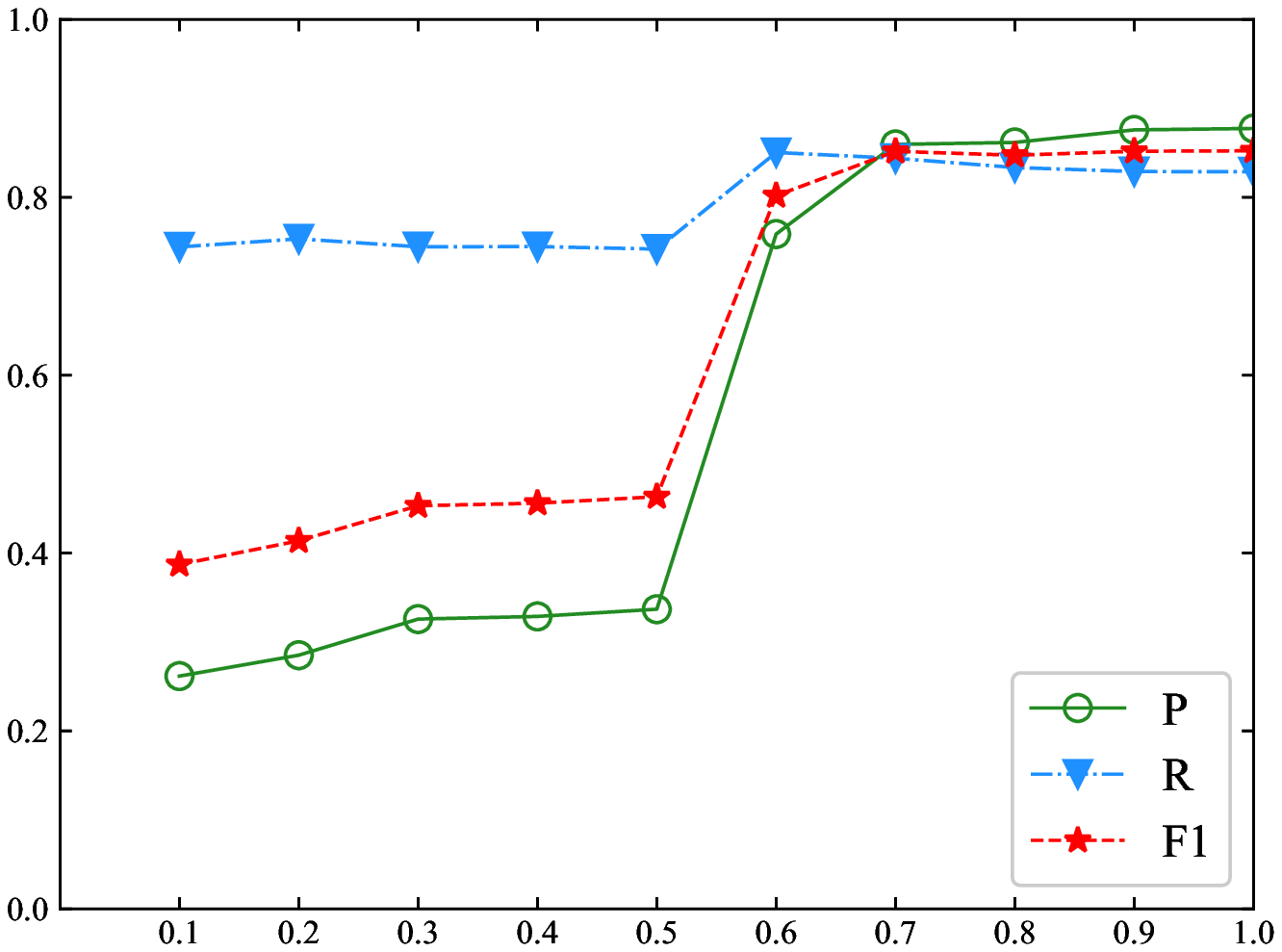}}
  \subfigure[IoU ($\gamma$) of BR$_{\text{int}}$]{\label{fig:IoU_for_20} \includegraphics[width=3in]{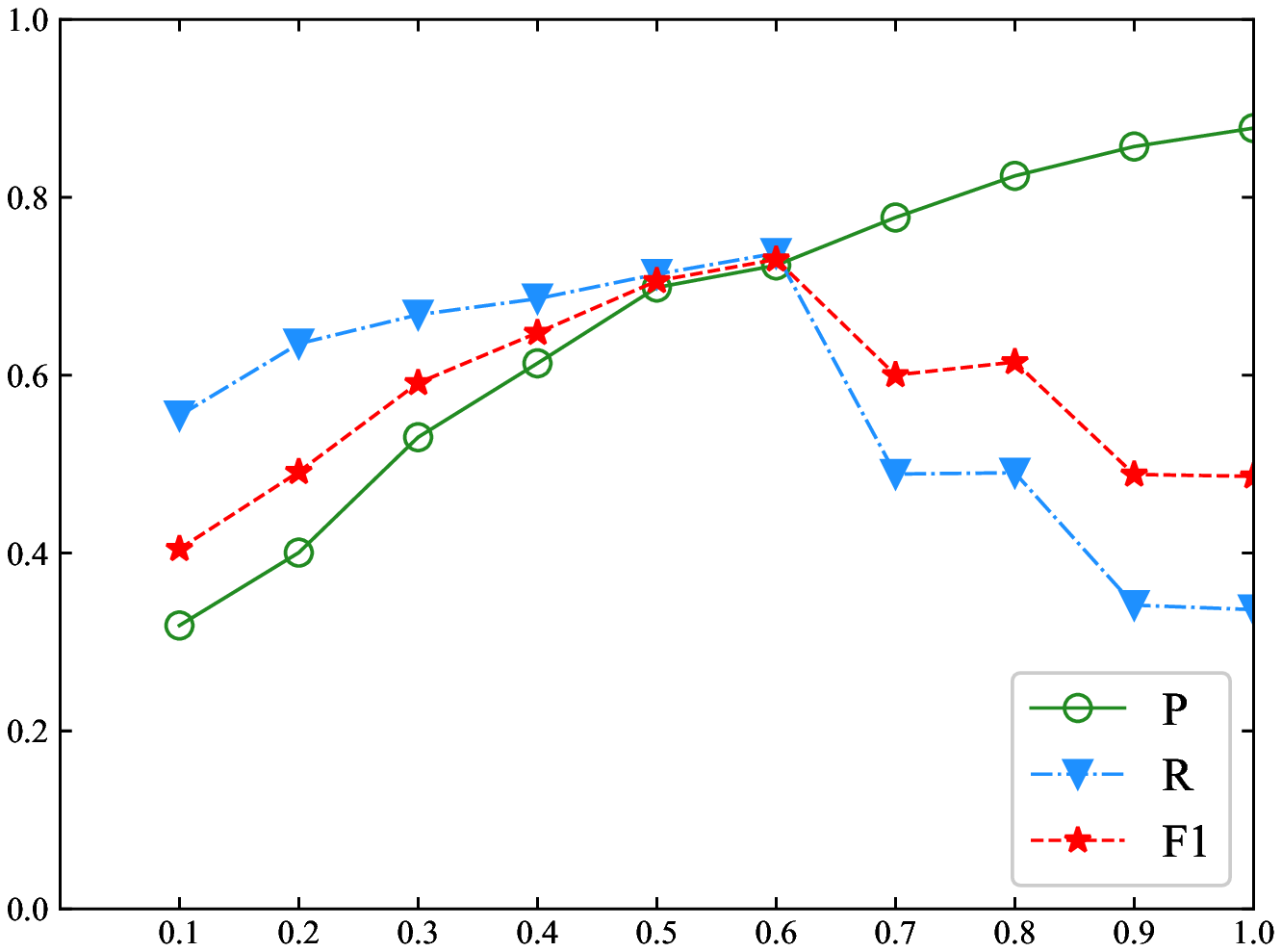}}
  \caption{Influence of IoU  Values}
  \label{fig:influence_iou}
\end{figure*}

\subsection{Influence of Model Parameters}

Because the IoU value and the NMS algorithm are influential on the BR model, in this section, we conduct experiments to analyse the influences of IoU and NMS.

\vspace{0.3cm}
\subsubsection{Influence of IoU} \
\label{tab:Influence_of_IoU}
\vspace{0.1cm}

In Equation \ref{equ:gamma}, a predefined parameter $\gamma$ is adopted to divide the training data into positive bounding box set $\mathbf{D}_p$ and negative bounding box set $\mathbf{D}_n$. Every bounding box in $\mathbf{D}_p$ has a high overlapping ratio with a true bounding box. The overlap enables each bounding box to contain semantic features about a truth box, which are used to train the linear layer. This is the key to supporting boundary regression. As Equation \ref{equ:loc_loss} shows, the location loss is computed from $\mathbf{D}_p$, which aggregates all position offsets between each bounding box and its relevant ground truth bounding box. Therefore, the IoU value directly determine the number of bounding boxes used for computing the location loss. 

This experiment is conduced to analyse the influence of IoU value $\gamma$ on the final performance. Because $\gamma=0.0$ cannot be used to collect positive bounding boxes,  the value is initialized from 0.1 to 1.0 with step size 0.1. The result is shown in Figure \ref{fig:influence_iou}.

In both the BR$_{\text{exh}}$ model and the BR$_{\text{int}}$ model, if $\gamma$ has a small value, $\mathbf{D}_p$ contains many bounding boxes with small overlapping ratios relevant to a true NE. In these bounding boxes, there are insufficient semantic features with respect to a true NE. The regression operation cannot guarantee appropriately learning of the location offset, which worsens the performance. The result indicates that a bounding box far from any truth box is less helpful for boundary regression.


The BR$_{\text{exh}}$ model achieves high performance when $\gamma$ is approximately 0.7. After $\gamma > 0.7$, the output of the BR$_{\text{exh}}$ model exhibits stable performance. The reason for the phenomenon is that, when the value of $\gamma$ is large enough, $\mathbf{D}_p$ contains almost exclusively enumerated ground truth boxes. As Equation \ref{equ:loc_loss} reveals, the influence of regression is weakened. On the other hand, because the BR$_{\text{exh}}$ model verifies every NE candidate with length from 1 to 6, in this condition, the BR$_{\text{exh}}$ model is almost degenerated into a traditional classification model. Its performance is heavily dependent upon the output of the softmax layer.

In the BR$_{\text{int}}$ model, the highest performance is achieved when $\gamma$ is approximately 0.6. When the value of $\gamma$ exceeds 0.6, the performance is considerably diminished. In the BR$_{\text{int}}$ model, a large $\gamma$ means that $\mathbf{D}_p$ contains a smaller number of positive bounding boxes. Their boundaries are almost precisely matched with ground truth boxes. In particular, when $\gamma=1.0$, $\mathbf{D}_p$ only contains grounding truth boxes. As Equation \ref{equ:loc_loss} shows, in the training process, the location loss is always zero. Therefore, the linear layer cannot be trained appropriately.


\vspace{0.3cm}
\subsubsection{Influence of NMS}\
\vspace{0.1cm}

\begin{figure*}[h]
  \centering 
  \subfigure[NMS ($\lambda$) of BR$_{\text{exh}}$]{\label{fig:NMS_for_6} \includegraphics[width=3in]{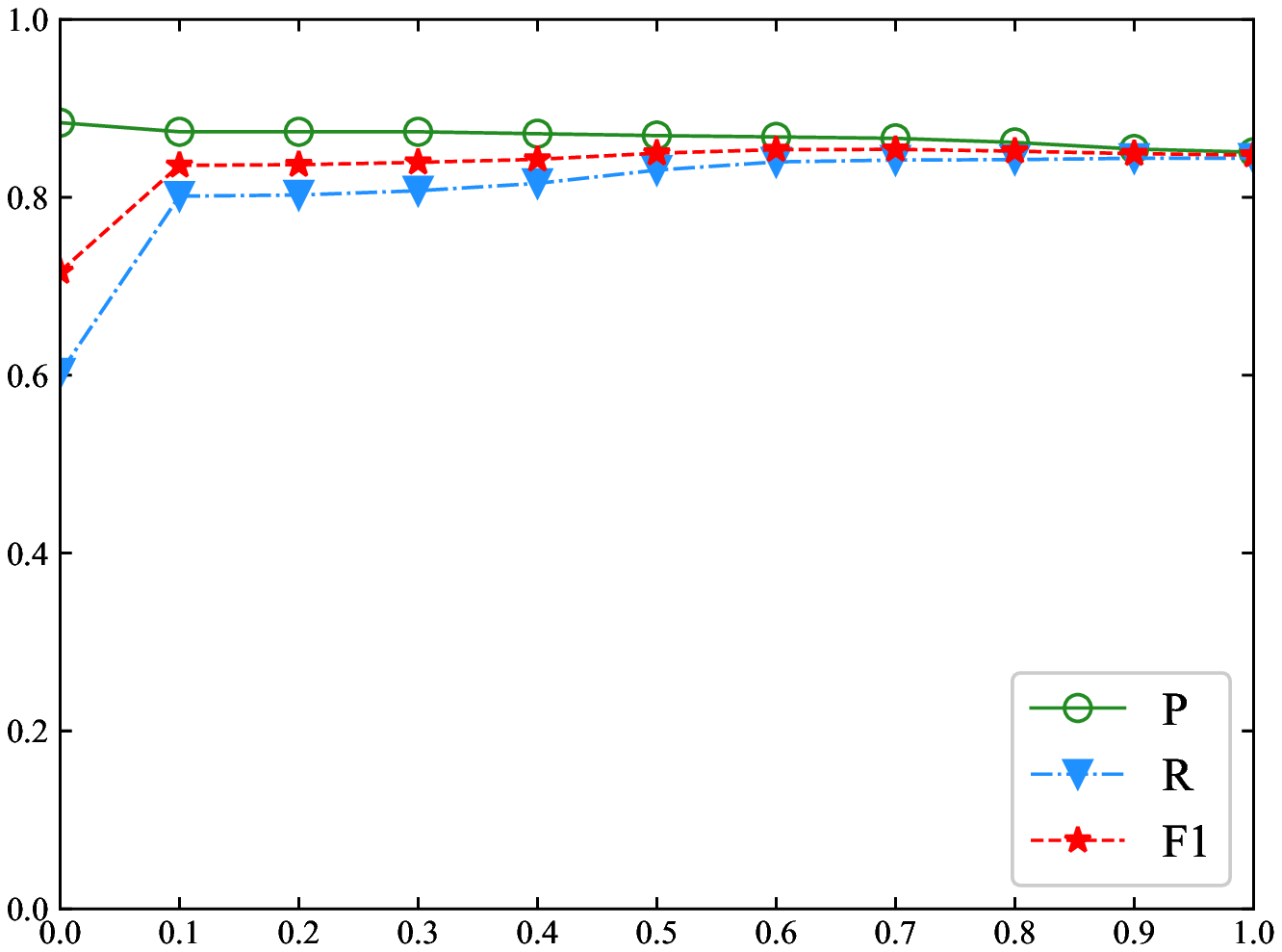}}
  \subfigure[NMS ($\lambda$) of BR$_{\text{int}}$]{\label{fig:NMS_for_20} \includegraphics[width=3in]{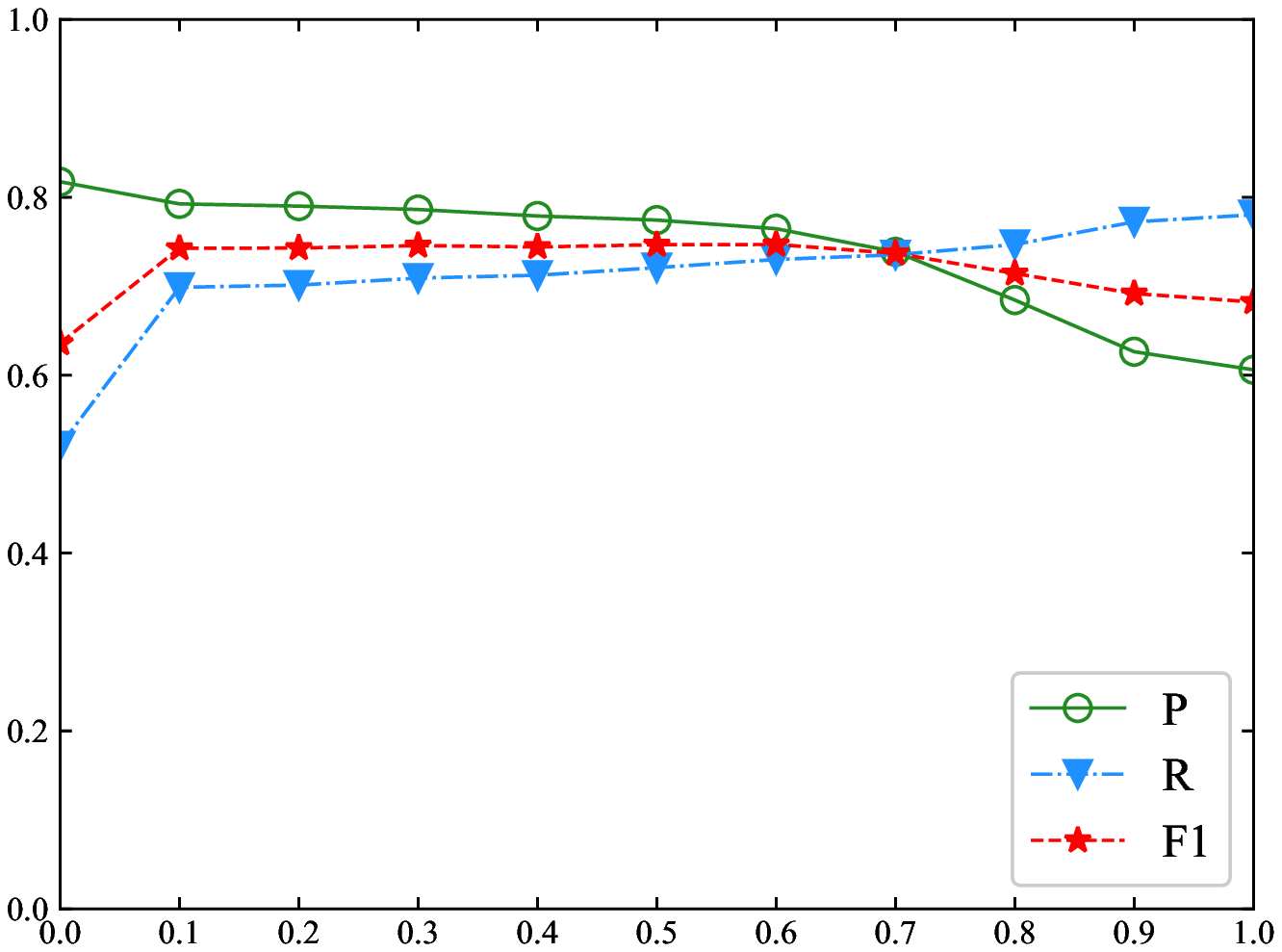}}
  \caption{Influence of NMS  Values}
  \label{fig:influence_nms}
\end{figure*}

In the testing process, the BR model adopts an one-dimensional NMS algorithm to select true bounding boxes from the output (as Table \ref{tab:NMS} shows). The NMS algorithm is originally designed to support object detection in computer vision, where a rectangle is adopted to frame an object. One difference between object detection and entity recognition is that, when detecting an object, a rectangle is permitted to have mutual overlap with the reference object. On the other hand, recognizing an NE requires that both start and end boundaries of an NE should be precisely matched. In this experiment, we study the influence of NMS on the nested NE recognition. The result is shown in Figure \ref{fig:influence_nms}.

The results show that $\lambda=0.0$ leads to the lowest recall because many bounding boxes are discarded. When $\lambda>0.1$, enlarging $\lambda$ slowly improves the performance. Because bounding boxes belonging to a true NE are closely overlapped, if the $\lambda$ is not large enough, enlarging $\lambda$ exerts little influence on the performance. Therefore, a stable performance is achieved when $\lambda$ takes value from 0.1 to 0.6. The BR models achieve the best performance around $\lambda=0.6$. Comparing the BR$_{\text{int}}$ model with the BR$_{\text{exh}}$ model, the performance of the BR$_{\text{int}}$ model is decreased when $\lambda > 0.6$. The reason for this is that the BR$_{\text{exh}}$ exhaustively enumerates all NE candidates with length up to 6. The output contains a large number of bounding boxes which have precisely matched boundaries.

In the NE recognition task, identifying an NE heavily depends on its contextual features. Therefore, highly overlapped bounding boxes may refer to different true NEs, which will be discarded when $\lambda \geq 0.6$. This problem can be avoided by setting $\lambda = 1.0$. In the BR model, the performance of $\lambda=1.0$ is the same as disabling the NMS algorithm. In this setting, only fully overlapped bounding boxes are considered and erased as redundant boxes. Because many bounding boxes are remained even they have high overlapping ratio, this setting has a higher recall. However, it worsens the precision. As shown in Figure \ref{fig:NMS_for_20}, $\lambda = 1.0$ leads to a poor F1 score. 
 
\subsection{Time Complexity of Boundary Regression}

In object detection of computer vision, compared with multistage pipeline models (e.g., R-CNN \cite{girshick2014rich}), an end-to-end framework is hailed due to the superiority of high speed (e.g., Faster R-NN \cite{ren2015faster}). The reason for this is that the background of an image is learned in a single pass in the training process. It is shared by all proposed regions in an image.

To show the time complexity of boundary regression, in this experiment, we compare our model with those of Zheng et al. \cite{zheng2019boundary} and Wang et al. \cite{wang2020pyramid}. Zheng et al. \cite{zheng2019boundary} present a boundary-aware neural model which detects entity boundaries. Boundary-relevant regions are then utilized to predict entity categorical labels. The boundary detection and region prediction share the same bidirectional LSTM for feature extraction. Wang et al. \cite{wang2020pyramid} present a pyramid-shaped model stacked with neural layers. This model directly implements NE span prediction. Therefore, it has a higher speed. 

\begin{figure}[htbp]
	\centering
	\includegraphics[width=7.5cm]{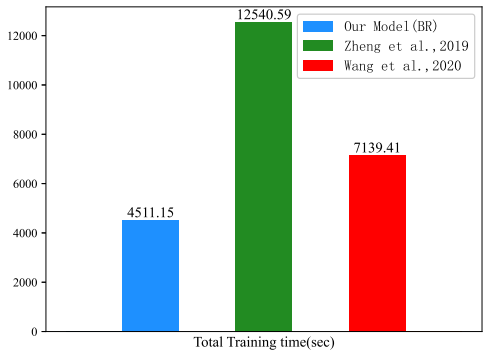}
	\caption{Time Complexity Comparison}
	\label{fig:time}
\end{figure}

In this experiment, we implement these models on the ACE English corpus with the same data split, settings and GPU platform. The times required to train these models are shown in Figure \ref{fig:time}, where the height of the histograms represents the time cost in seconds.

In comparison with them, boundary regression achieves the least time complexity. The BR model has two characteristics which support high speed recognition. First, feature maps are generated from a basic network. They are shared by all bounding boxes in a sentence. In fact, all bounding boxes are mutually overlapped. They are parts of feature maps, which considerably reduce model parameters. Second, every bounding box has location parameters. Therefore, in the learning process, the IoU value can be adopted to filter negative bounding boxes. This strategy is effective to reduce the time complexity.

\subsection{Visualization of Boundary Regression}

For a better understanding of boundary regression and investigating more details of the BR model, in the follows, we present a visualization about boundary regression.

\begin{figure*}[h]
  \centering 
  \subfigure[Iterations: 0]{\label{fig:re_a} \includegraphics[width=3in]{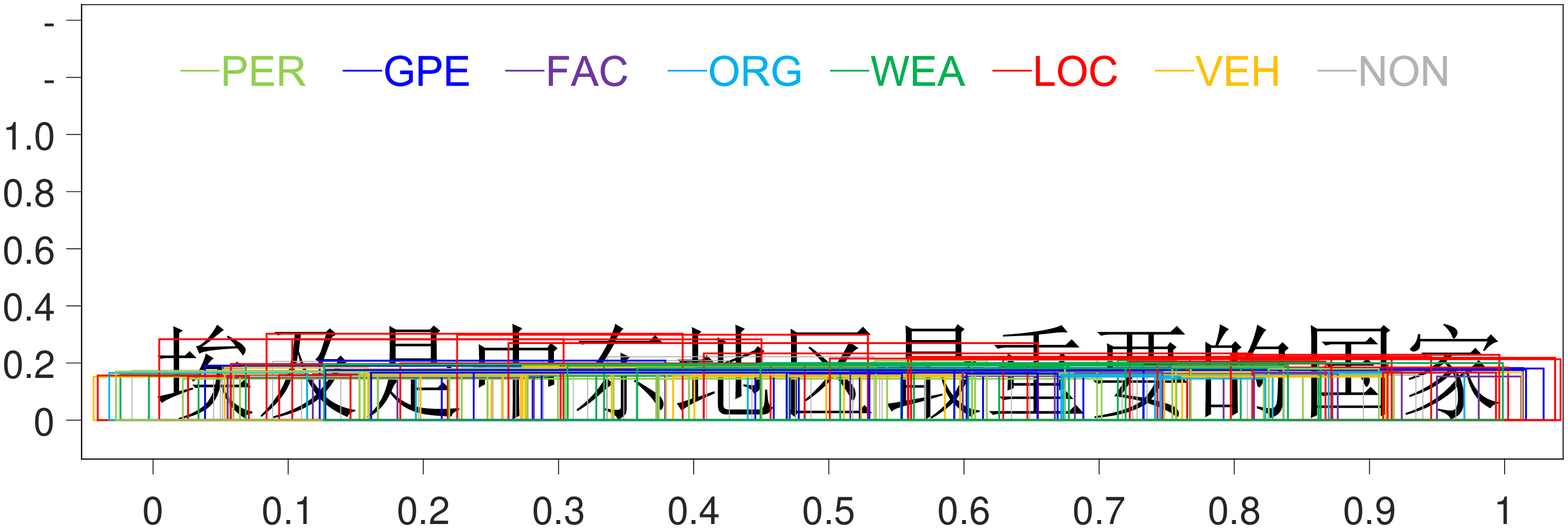}}
  \subfigure[Iterations: 1]{\label{fig:re_b} \includegraphics[width=3in]{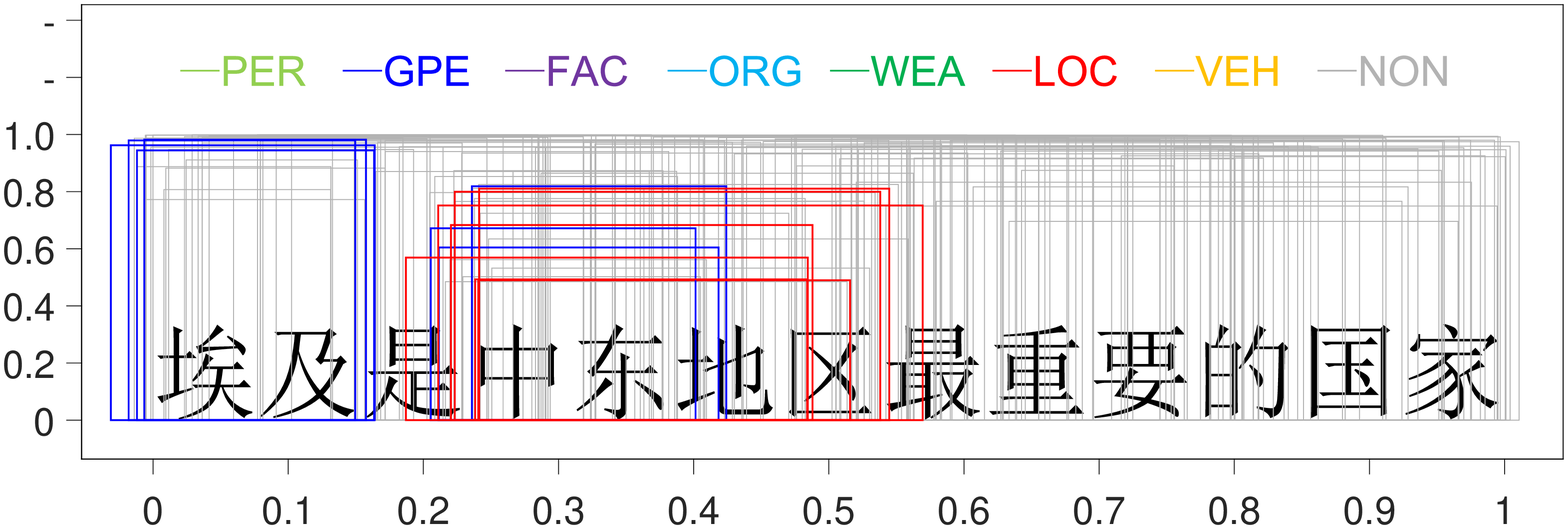}}
   \subfigure[Iterations: 10]{\label{fig:re_c} \includegraphics[width=3in]{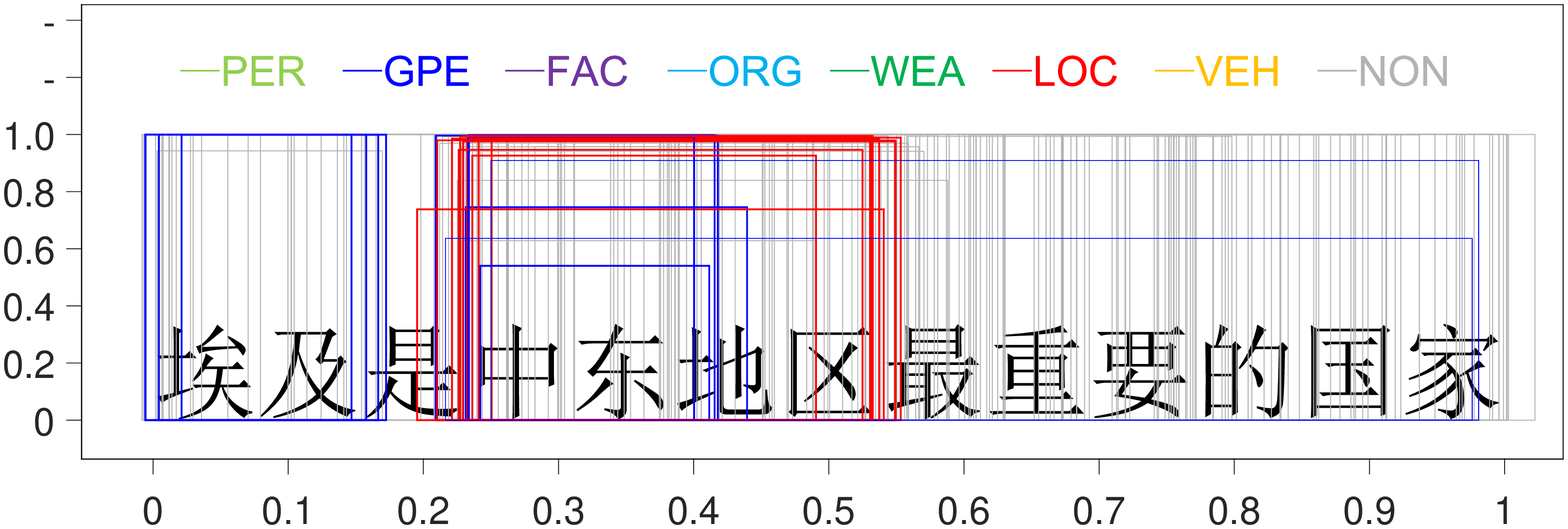}}
  \subfigure[Iterations: 100]{\label{fig:re_d} \includegraphics[width=3in]{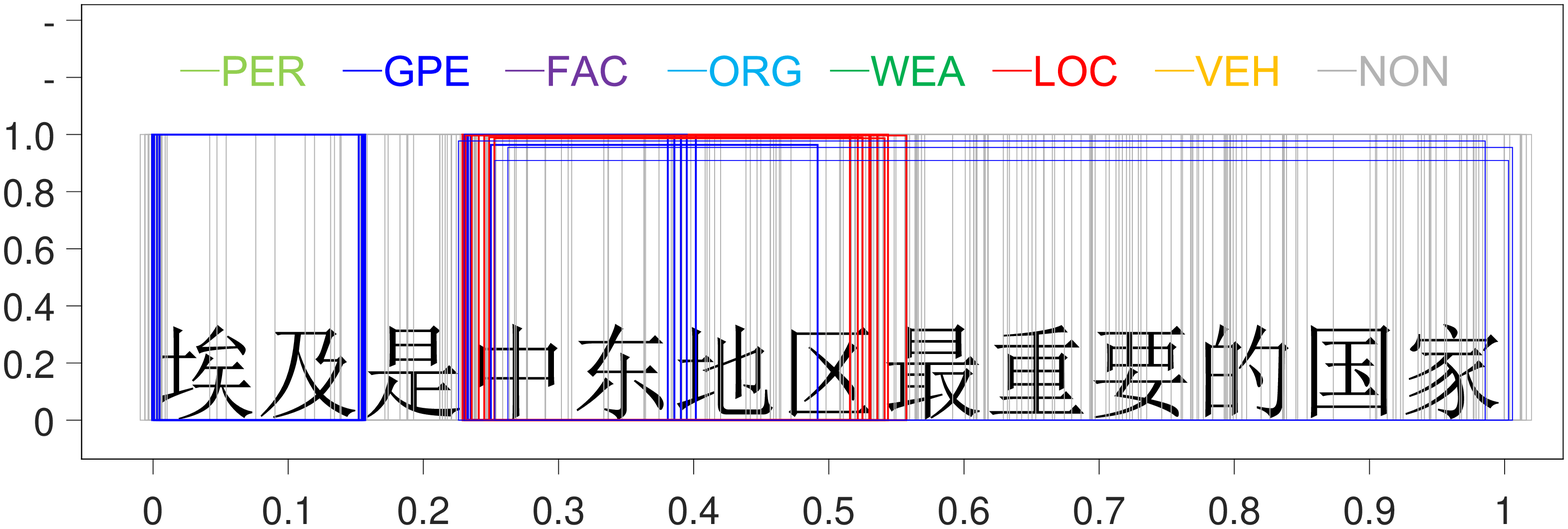}}
  \subfigure[Iterations: 200]{\label{fig:re_e} \includegraphics[width=3in]{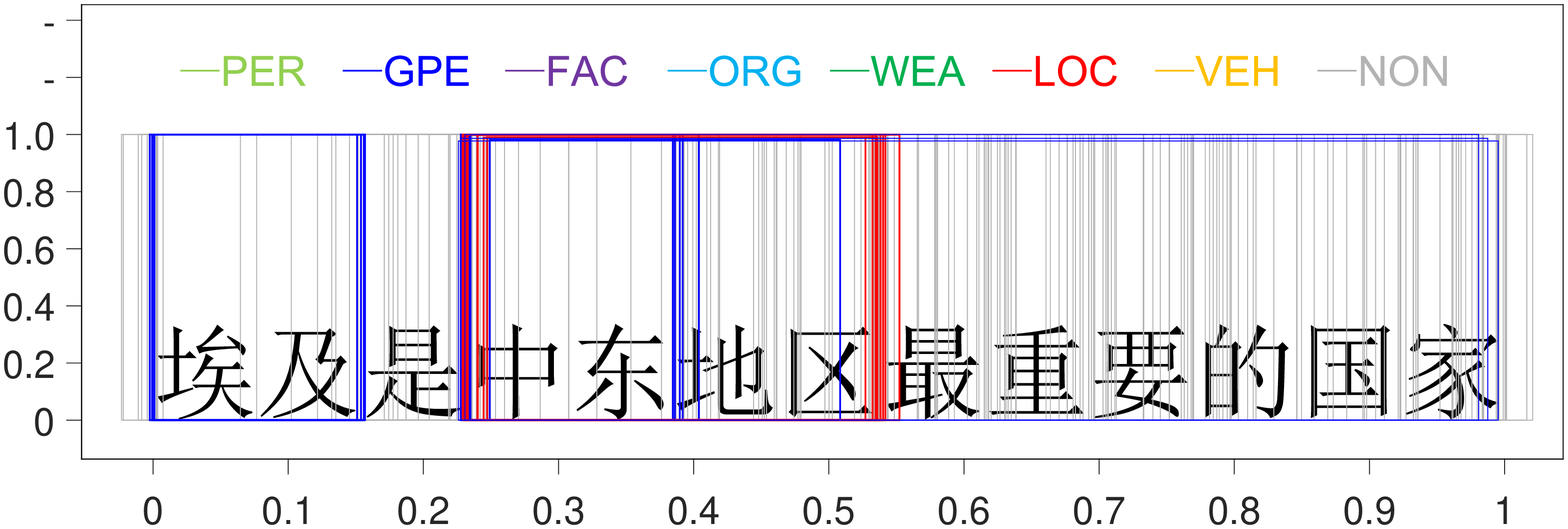}}
    \subfigure[Iterations: 300]{\label{fig:re_f} \includegraphics[width=3in]{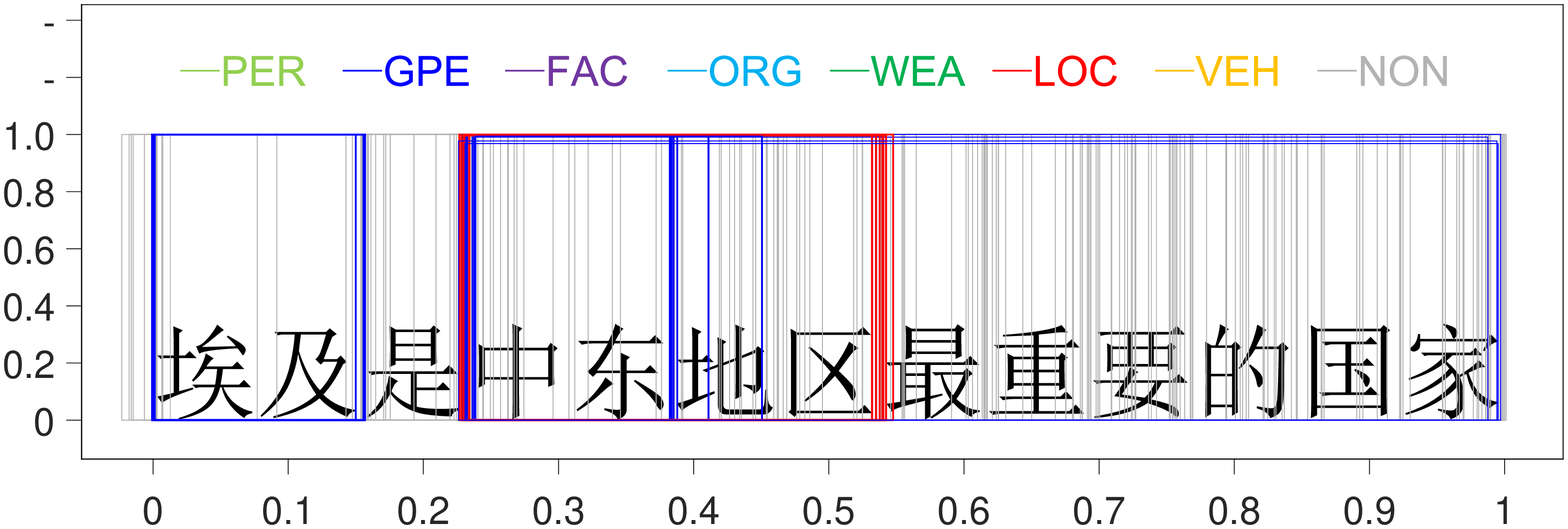}}
\caption{ Visualization of the Bounding Box Regression}
  \label{fig:textural_regression}
\end{figure*}

A sentence ``\begin{CJK}{UTF8}{gbsn}埃及是中东地区最重要的国家\end{CJK}''\footnote{It can be translated as: ``Egypt is the most important country in the Middle East area''.} is selected from the testing data. It contains four nested NEs: ``\begin{CJK}{UTF8}{gbsn}埃及\end{CJK}'' (Egypt, GPE), ``\begin{CJK}{UTF8}{gbsn}中东地区最重要的国家\end{CJK}'' (the most important country in the Middle East area, GPE), ``\begin{CJK}{UTF8}{gbsn}中东地区\end{CJK}'' (the Middle East area, LOC), and ``\begin{CJK}{UTF8}{gbsn}中东\end{CJK}'' (the Middle East, GPE). A bounding box is denoted by 3 parameters $s_i$, $l_i$ and $c_i$, which represent the starting position of the box, the length of the box and the class probability of the box, respectively. To visualize bounding boxes, a bounding box is drawn as a rectangle. The horizontal ordinate represents the boundary positions of the bounding boxes in a sentence, which are normalized to $[0,1]$. The vertical coordinate represents the classification confidence score. The colours of the bounding box represent NE types. To generate bounding boxes, the selected sentence is predicted by a pre-trained BR model. All outputted bounding boxes are collected and drawn with respect to the sentence. The result is shown in Figure \ref{fig:textural_regression}.

In Figure \ref{fig:re_a}, bounding boxes were predicted by the BR model without training (0 iterations). From Figure \ref{fig:re_b} to Figure \ref{fig:re_f}, the BR model is trained  with different rounds (denoted in the titles of subfigures). Because the regression operation may output negative values for parameters $s_i$ and $l_i$, we filter bounding boxes with $l_i \leq 0$ or $s_i+l_i > 1$ (beyond the sentence range). 

In Figure \ref{fig:re_a}, there is no tendency between bounding boxes. They are distributed evenly across the whole sentence and all NE types. In Figure \ref{fig:re_b}, the BR model is implemented on the training data in only one round. One interesting phenomenon is that red bounding boxes and blue bounding boxes are grouped around NEs quickly. Furthermore, other true entity types are appropriately depressed. From Figure \ref{fig:re_c} to Figure \ref{fig:re_f}, when the number of iterations is increased, there are two tendencies in bounding boxes. First, the BR model becomes more confident in the entity type prediction, which increases the classification confidence of the bounding boxes. Second, the locations of bounding boxes are approaching the true NEs. This indicates that the regression operation to locate NEs is feasible.

Overlapped bounding boxes are the key to solving the nested NE problem. Figure \ref{fig:re_f} shows that nested NEs are distinguished appropriately. We have tracked several bounding boxes and found that bounding boxes do not approach true NEs smoothly and directly. There are some oscillations between them. In the training process, a bounding box may match the true NE perfectly, while moving away in the next iteration. However, in accordance with the increase in the number of training steps, the oscillation tends toward stability. 

\section{Related Work}
\label{sec:related_work}

Because the BR model is motivated by object detection from computer vision, in the following, we divide the related work into two parts: object detection and NE recognition.

Object detection is implemented in a multistage pipeline in the early stage. A typical object detection model is often composed of three stages: segmentation, feature extraction and classification. Segmentation is implemented to generate possible object locations for prediction. Generic algorithms (e.g., selective search) are often adopted to avoid exhaustive searching. Feature extraction is implemented to extract higher-order abstract features from raw input images. The output of this process is often denoted as feature maps. The feature extraction process can be encapsulated as a basic network truncated from a standard architecture for high-quality image classification, including the VGG-16 network \cite{simonyan2014very}, GoogLeNet \cite{szegedy2015going}, etc. Finally, an output layer (e.g., a linear SVM or a softmax layer) is used to predict confidence scores for each proposed region.

End-to-end object detection models can be optimized globally and share computation between inputs. These models are often similar in the feature extraction layer and output layer, where a basic network is adopted to generate {\em conv} feature maps, and two fully connected layers synchronously output class probabilities and object locations. The main difference is the strategy to generate the region proposal. For example, Faster R-NN adopts anchor boxes to generate region proposals per feature map \cite{ren2015faster}. Erhan et al. \cite{erhan2014scalable} use a single deep neural network to generate a small number of bounding boxes. Redomon et al. \cite{redmon2016you} divide an image into grids associated with a number of bounding boxes. Liu et al. \cite{liu2016ssd} use a basic network that maps an image into multiple feature maps for generating default boxes with different aspect ratios and scales.

In the field of NE recognition, neural networks have also received great attention. Early models usually adopt a sequence model to output flattened NEs (e.g., LSTM, Bi-LSTM or Bi-LSTM-CNN). To handle the nesting problem, the sequence model is redesigned. It has three variants: the layering, cascading and joint models \cite{alex2007recognising}. Parsing trees are also widely used to represent nested NEs in a tree structure~\cite{finkel2009nested}. For example, Finkel et al. \cite{finkel2009joint} use internal and structural information of parsing trees to flatten nested NEs. Zhang et al. \cite{zhang2014parsing} adopted a transition-based parser. Jie et al. \cite{jie2017efficient} tried to capture the global dependency of a parsing tree. 

Recently, many models have been designed to recognize nested NEs directly. Lu et al. \cite{lu2015joint} resolve nested NEs into a hypergraph representation. Xu et al. \cite{xu2016fofe} and Sohrab et al.~\cite{sohrab2018deep} verify every possible fragment up to a certain length. Wang et al.~\cite{wang2018neural} map a sentence with nested mentions to a designated forest. Ju et al.~\cite{ju2018neural} proposed an iterative method that implements a sequence model in the output of a previous model.  Lin et al. \cite{lin2019sequence} propose a head-driven structure. Li et al. \cite{li2017recognizing} combined outputs of a Bi-LSTM-CRF network with another Bi-LSTM network. Strakova et al.~\cite{strakova2019neural} proposed a sequence-to-sequence model. Zheng et al. \cite{zheng2019boundary} proposed an end-to-end boundary-aware neural model. In Chen et al. \cite{chen2015boundary}, a boundary assembling (BA) model is designed to recognize nested NEs. The BA model identifies NE boundaries, assembles them into NE candidates, and picks the most likely ones. For a broad understanding of the NER problem, the interested reader can refer to the survey paper \cite{li2020survey} for deep neural network based NE recognition.

\section{Conclusion and Future Work}
\label{sec:future_work}

In this paper, we proposed a boundary regression model for nested NE recognition. The BR model can be seen as a framework to support nested NE recognition. In Section \ref{sec:feature_map}, we divide the BR model into two modules:  ``perceptional module'' and ``cognitive module''.  In the perceptional module'', various deep architectures can be designed to extract high order abstract features from raw inputs. In cognitive module, instead of bounding boxes, abstract NE representations can be defined with other position and shape parameters. These issues are left as our future work. For enumerating NE candidates, new strategies can be designed to support region proposal. These issues are left as our future work. They are also open for researchers who are interested in this work.

\section{Acknowledgment}
This work is supported by the Joint Funds of the National Natural Science Foundation of China (Nos. 62166007, 62050194, and 62037001, 62066007, 62066008, 61721002), the National Natural Science Foundation of China under Grant No. U1836205, and the Key Projects of Science and Technology of Guizhou Province under Grant No. [2020]1Z055.

\bibliography{reference}
\end{document}